\documentclass[lettersize,journal]{IEEEtran}
\usepackage{url}
\usepackage{verbatim}
\usepackage{graphicx}
\usepackage{color}
\usepackage{caption}

\usepackage{moreverb,url}
\usepackage[colorlinks,bookmarksopen,bookmarksnumbered,citecolor=black,urlcolor=black]{hyperref}
\usepackage{stackengine}
\usepackage{amsfonts,amssymb,amsmath,bm, mathabx}
\usepackage[ruled,vlined]{algorithm2e}
\usepackage{multirow}
\usepackage{tabularx}
\usepackage{tabularray}
\UseTblrLibrary{siunitx}

\usepackage{xcolor}

\definecolor{sampler}{RGB}{209, 191, 196}
\definecolor{scalar}{RGB}{193, 197, 207}
\definecolor{vector}{RGB}{194, 204, 198}
\definecolor{logits}{RGB}{209, 198, 210}

\definecolor{redhak}{RGB}{231, 87, 83}
\definecolor{bluehak}{RGB}{5, 148, 177}
\definecolor{yellowhak}{RGB}{133, 165, 17}
\definecolor{orangehak}{RGB}{242, 142, 44}
\definecolor{greenhak}{RGB}{133, 165, 17}
\definecolor{purplehak}{RGB}{98, 84, 254}

\usepackage[]{glossaries}
\newacronym{dgm}{DGM}{Deep Generative Models}

\newacronym{gail}{GAIL}{Generative Adversarial Imitation Learning}
\newacronym{sqil}{SQIL}{Soft-Q Imitation Learning}

\newacronym{iflow}{iFlows}{ImitationFlows}
\newacronym{cep}{CEP}{Composable Energy Policies}
\newacronym{se3dif}{SE(3)-DiF}{SE(3)-DiffusionFields} 

\newacronym{nlp}{NLP}{Natural Language Processing}
\newacronym{cv}{CV}{Computer Vision}

\newacronym{gp}{GP}{Gaussian Process}
\newacronym{gan}{GAN}{Generative Adversarial Networks}
\newacronym{vae}{VAE}{Variational Autoencoders}
\newacronym{cvae}{cVAE}{Conditional Variational Autoencoders}
\newacronym{ebm}{EBM}{Energy Based Models}
\newacronym{sbm}{SBM}{Score based Models}
\newacronym{nf}{NFlow}{Normalizing Flows}

\newacronym{dm}{DM}{Diffusion Models}
\newacronym{ddpm}{DDPM}{Denoising Diffusion Probabilistic Models}
\newacronym{ncsn}{NCSN}{Noise Conditioned Score Network}
\newacronym{smld}{SMLD}{Score Matching with Langevin Dynamics}

\newacronym{map}{MAP}{Maximum a Posteriori}

\newacronym{mdm}{MDM}{Mixture Density Models}
\newacronym{gmm}{GMM}{Gaussian Mixture Models}

\newacronym{nerf}{NeRF}{Neural Radiance Fields}

\newacronym{bc}{BC}{Behavioural Cloning}
\newacronym{il}{IL}{Imitation Learning}
\newacronym{irl}{IRL}{Inverse Reinforcement Learning}
\newacronym{ioc}{IOC}{Inverse Optimal Control}
\newacronym{lfd}{LfD}{Learning from Demonstration}
\newacronym{em}{EM}{Expectation Maximization}
\newacronym{promp}{ProMP}{Probabilistic Movement Primitives}
\newacronym{dmp}{DMP}{Dynamic Movement Primitives}
\newacronym{seds}{SEDS}{Stable Estimator of Dynamical Systems}
\newacronym{gmr}{GMR}{Gaussian Mixture Regressor}
\newacronym{gpr}{GPR}{Gaussian Process Regressor}
\newacronym{lwr}{LWR}{Locally Weighted Regressor}
\newacronym{kmp}{KMP}{Kernelized Movement Primitives}
\newacronym{clf}{CLF}{Control Lyapunov Function}
\newacronym{wsaqf}{WSAQF}{Weighted Sum of Asymmetric Quadratic Function)}
\newacronym{nilc}{NILC}{Neurally Imprinted Lyapunov
Candidate}
\newacronym{clfdm}{CLF-DM}{Control Lyapunov Function-based Dynamic Movements}
\newacronym{cnmp}{CNMP}{Conditional Neural Movement Primitives}
\newacronym{tpgmm}{TP-GMM}{Task Parameterized GMM}

\newacronym{mp}{MP}{Movement Primitive}
\newacronym{mpflows}{MPFlows}{Movement Primitive Flows}

\newacronym{gcl}{GCL}{Guided Cost Learning}

\newacronym{mle}{MLE}{Maximun Likelihood Estimation}
\newacronym{sde}{SDE}{Stochastic Differential Equation}
\newacronym{ode}{ODE}{Ordinary Differential Equation}

\newacronym{probs}{ProbS}{Probabilistic Segmentation}
\newacronym{crf}{CRF}{Conditional Random Fields}
\newacronym{ppca}{PPCA}{Probabilistic Principal Component Analysis}
\newacronym{gmcc}{GMCC}{Generalized Multiple Correlation Coeficcient}
\newacronym{hri}{HRI}{Human-Robot Interaction}
\newacronym{ip}{IP}{Interaction Primitives}
\newacronym{hmm}{HMM}{Hidden Markov Model}
\newacronym{cac}{CAC}{Canonical Correlation Coefficient}
\newacronym{rv}{$R_v$}{$R_v$ Coefficient}
\newacronym{dcor}{dCor}{Distance Correlation}
\newacronym{dtw}{DTW}{Dynamic Time Warping}
\newacronym{edr}{EDR}{Edit Distance With Real Penalty}
\newacronym{twed}{TWED}{Time Warp Edit Distance}
\newacronym{r2}{$R^2$}{Coefficient of Determination}
\newacronym{sqp}{SQP}{Successive Quadratic Programming}
\newacronym{rkhs}{RKHS}{Reproducing Kernel Hilbert Space}
\newacronym{icnn}{ICNN}{Input-Convex Neural Network}

\newacronym{pca}{PCA}{Principal Component Analysis}

\newacronym{maf}{MAF}{Masked Autoregressive Flow}
\newacronym{iaf}{IAF}{Inverse Autoregressive Flow}
\newacronym{node}{N-ODE}{Neural ODE}
\newacronym{nsflow}{NSF}{Neural Spline Flows}
\newacronym{cnf}{CNF}{Conditional Normalizing Flows}
\newacronym{ffjord}{FFJORD}{Free-form Jacobian of Reversible Dynamics}

\newacronym{inn}{INN}{Invertible Neural Networks}
\newacronym{mcmc}{MCMC}{Markov Chain Monte Carlo}
\newacronym{ld}{LD}{Langevin Dynamics}

\newacronym{cd}{CD}{Contrastive Divergence}
\newacronym{nce}{NCE}{Noise Contrastive Estimation}
\newacronym{ce}{CE}{Cross-Entropy}
\newacronym{dsm}{DSM}{Denoising Score Matching}

\newacronym{ik}{IK}{Inverse Kinematics}

\newacronym{sdf}{SDF}{Signed Distance Field}
\newacronym{deepsdf}{DeepSDF}{Deep Signed Distace Field}

\newacronym{emd}{EMD}{Earth Mover Distance}

\newacronym{mlp}{MLP}{Multi Layer Perceptron}
\newacronym{rrt}{RRT}{Rapidly-exploring Random Trees}

\newacronym{relu}{ReLU}{Rectified Linear Unit}

\newacronym{vn}{VN}{Vector Neuron}

\newacronym{icp}{ICP}{Iterative Closest Point}

\newacronym{rmp}{RMP}{Riemannian Motion Policies}
\newacronym{mpc}{MPC}{Model Predictive Control}

\newacronym{svi}{SVI}{Structured Variational Inference}
\newacronym{vi}{VI}{Variational Inference}
\newacronym{hrl}{HRL}{Hierarchical Reinforcement Learning}
\newacronym{apf}{APF}{Artificial Potential Fields}

\newacronym{reps}{REPS}{Relative Entropy Policy Search}
\newacronym{rl}{RL}{Reinforcement Learning}

\newacronym{lmdp}{LMDP}{linearly-solvable Markov Decision Processes}
\newacronym{dwa}{DWA}{Dynamic Window Approach}
\newacronym{svf}{SVF}{Stable Vector Fields}

\newacronym{mse}{MSE}{Mean Squared Error}

\newacronym{stomp}{STOMP}{Stochastic Trajectory Optimization for Motion Planning}
\newacronym{gpmp}{GPMP}{Gaussian Process Motion Planning}
\newacronym{chomp}{CHOMP}{Covariant Hamiltonian Optimization for Motion Planning}

\newacronym{llm}{LLM}{Large Language Models}

\def\1{\bm{1}}

\def\RR{\mathbb{R}}
\def\d{{\textrm{d}}}

\def\E{{\mathbb{E}}}

\def\vzero{{\bm{0}}}

\def\vmu{{\bm{\mu}}}
\def\vtheta{{\bm{\theta}}}
\def\va{{\bm{a}}}

\def\vc{{\bm{c}}}

\def\ve{{\bm{e}}}

\def\vg{{\bm{g}}}

\def\vo{{\bm{o}}}
\def\vp{{\bm{p}}}
\def\vq{{\bm{q}}}

\def\vs{{\bm{s}}}

\def\vx{{\bm{x}}}

\def\vz{{\bm{z}}}

\def\vmu{{\boldsymbol{\mu}}}
\def\vpsi{{\boldsymbol{\psi}}}
\def\vphi{{\boldsymbol{\phi}}}
\def\vsigma{{\boldsymbol{\sigma}}}
\def\veta{{\boldsymbol{\eta}}}
\def\vtheta{{\boldsymbol{\theta}}}
\def\vtau{{\boldsymbol{\tau}}}

\def\mC{{\bm{C}}}
\def\mD{{\bm{D}}}
\def\mE{{\bm{E}}}

\def\mH{{\bm{H}}}
\def\mI{{\bm{I}}}
\def\mJ{{\bm{J}}}

\def\mM{{\bm{M}}}

\def\mP{{\bm{P}}}

\def\mS{{\bm{S}}}

\DeclareMathAlphabet{\mathsfit}{\encodingdefault}{\sfdefault}{m}{sl}
\SetMathAlphabet{\mathsfit}{bold}{\encodingdefault}{\sfdefault}{bx}{n}

\def\gD{{\mathcal{D}}}
\def\gE{{\mathcal{E}}}

\def\gJ{{\mathcal{J}}}

\def\gL{{\mathcal{L}}}

\def\gN{{\mathcal{N}}}

\def\sD{{\mathbb{D}}}

\DeclareMathOperator*{\argmin}{arg\,min}

\usepackage{cleveref}

\usepackage{hyperref}
\hypersetup{
 colorlinks=teal,
 linkcolor=teal,
 filecolor=teal,
 citecolor=teal,      
 urlcolor=teal,
 }

\newcommand{\xxnote}[3]{}
\renewcommand{\xxnote}[3]{\color{#2}{#1: #3}}

\makeatletter
\let\@oldmaketitle\@maketitle%
\renewcommand{\@maketitle}{\@oldmaketitle%
    \vspace{5.0mm}
    \setcounter{figure}{0} 
    \centering
    \includegraphics[width=1.0\linewidth]{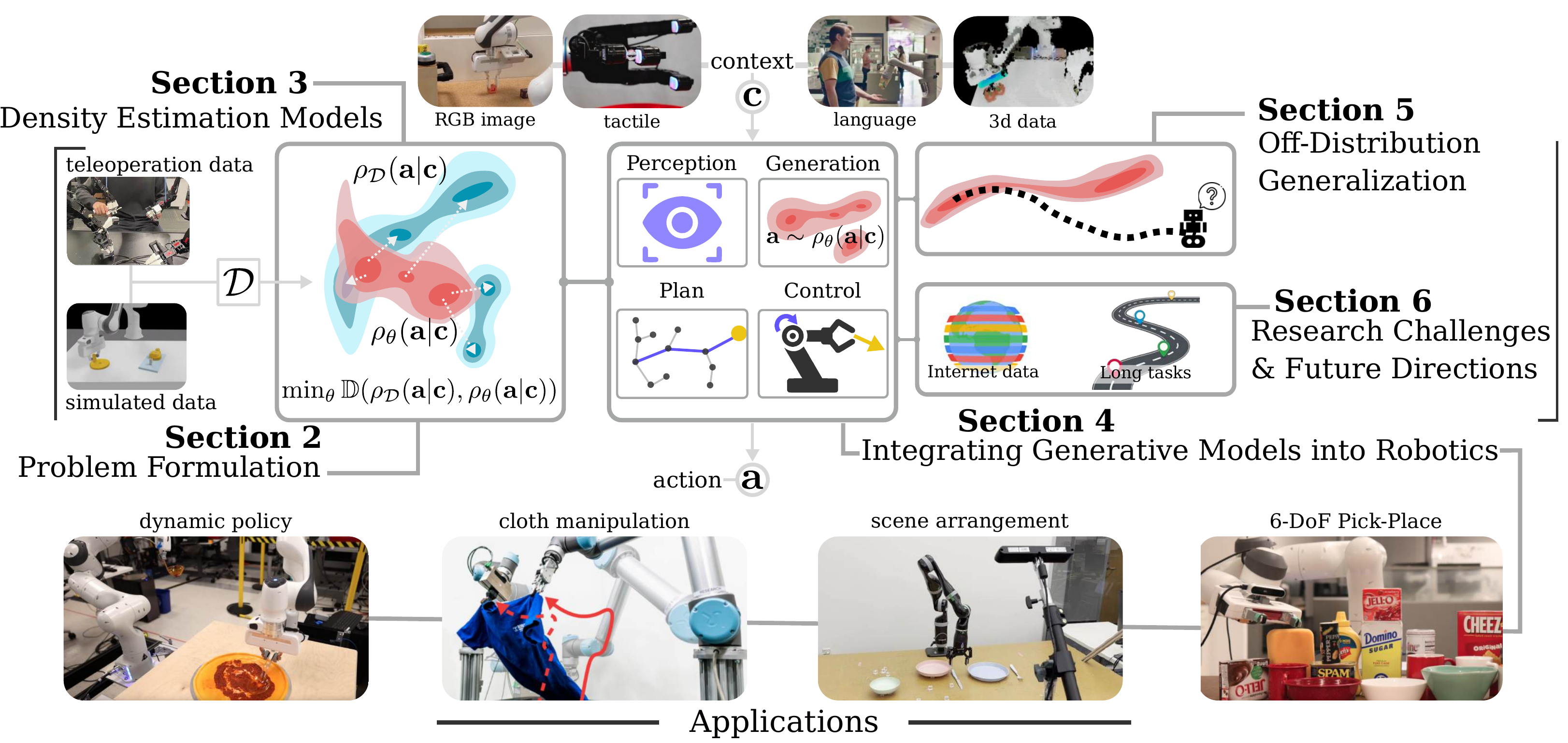}
    \captionof{figure}{Structure of the survey with references to the sections.}
    \label{fig:main}
}
\makeatother

\begin{document}

\title{Deep Generative Models in Robotics: \\A Survey on Learning from Multimodal Demonstrations}

\author{Julen Urain$^{1}$, Ajay Mandlekar$^{2}$, Yilun Du$^{3}$, Nur Muhammad Mahi Shafiullah$^{4}$, Danfei Xu$^{52}$, Katerina Fragkiadaki$^{6}$, Georgia Chalvatzaki$^{7}$, Jan Peters$^{718}$
\thanks{$^{1}$German Research Center for AI, $^2$Nvidia, $^3$Massachusetts Institute of Technology, $^4$New York University, $^5$Georgia Institute of Technology, $^6$Carnegie Mellon University, $^7$Technische Universität Darmstadt, $^8$Hessian.AI}
}

\markboth{}{}

\IEEEpubid{}

\maketitle

\begin{abstract}
Learning from Demonstrations, the field that proposes to learn robot behavior models from data, is gaining popularity with the emergence of deep generative models. 
Although the problem has been studied for years under names such as Imitation Learning, Behavioral Cloning, or Inverse Reinforcement Learning, classical methods have relied on models that don't capture complex data distributions well or don't scale well to large numbers of demonstrations.
In recent years, the robot learning community has shown increasing interest in using deep generative models to capture the complexity of large datasets.
In this survey, we aim to provide a unified and comprehensive review of the last year's progress in the use of deep generative models in robotics.
We present the different types of models that the community has explored, such as energy-based models, diffusion models, action value maps, or generative adversarial networks.
We also present the different types of applications in which deep generative models have been used, from grasp generation to trajectory generation or cost learning.
One of the most important elements of generative models is the generalization out of distributions. In our survey, we review the different decisions the community has made to improve the generalization of the learned models.
Finally, we highlight the research challenges and propose a number of future directions for learning deep generative models in robotics.
\end{abstract}


\section{Introduction}

\textbf{\gls{lfd}}~\cite{pomerleau1988alvinn, schaal1997learning}, also known as Imitation Learning~\cite{osa2018algorithmic, schaal1999imitation}, is the field that proposes to learn the desired robot behavior by observing and imitating a set of expert demonstrations.
Conditioned on observations of the scene and the desired task to be solved, the model, \textit{commonly known as policy}, is trained to generate actions that emulate the behavior in the expert demonstrations.
Depending on the task, these actions may represent desirable end-effector poses~\cite{mousavian20196, shridhar2023perceiver}, robot trajectories~\cite{chi2023diffusion, reuss2023goal} or desirable scene arrangements~\cite{liu2022structdiffusion, simeonov2023shelving}, to name a few.

\gls{lfd} includes several approaches to tackle this problem.
\textbf{\gls{bc}} methods~\cite{pomerleau1988alvinn} fit a conditional generative model to the actions conditioned on the observations.
Despite its shortcomings in sequential decision-making problems (e.g., compounding errors leading to covariate shift~\cite{ross2011reduction}), in practice, it has shown some of the most impressive results~\cite{shridhar2023perceiver, driess2023palm, chi2023diffusion, zhao2023learning} in part due to its stable and efficient training algorithms.

Alternatively, \textbf{\gls{irl}} \cite{ziebart2008maximum, fu2017learning} or variations such as \cite{ho2016generative, torabi2018generative} combine the demonstrations with trial-and-error in the environment~(i.e. \gls{rl}), resulting in policies that are more robust than \gls{bc}, but limited by less stable training algorithms.
Unlike \gls{bc}, which directly mimics the actions from the demonstrations, \gls{irl} focuses on inferring the underlying reward functions that the demonstrated behaviors aim to optimize, and applies \gls{rl} to infer the policy.
A key advantage of \gls{irl} is its ability to learn from mere observations~\cite{sermanet2018time, torabi2019recent}, without explicit information about the actions taken during the demonstrations.

In \gls{lfd}, the inherent characteristics of the demonstrations pose significant challenges.
Typically, the collected data is suboptimal, noisy, conditioned on high-dimensional observations, and includes multiple modes of behavior~\cite{mandlekar2018roboturk, mandlekar2019scaling, padalkar2023open}. This diversity can be observed in the multiple ways to grasp a given object, the preferences of the experts in providing the demonstrations, or the divergences between experts.
These inherent properties of the data lead the researchers to find models that can properly capture its distribution.

Traditionally, before deep learning became standard, \gls{lfd} methods often used \gls{gp}~\cite{grochow2004style, shon2005robotic}, \gls{hmm}~\cite{tso1996hidden, yang1997human}, or \gls{gmm}~\cite{calinon2007learning} to represent the generative models.
However, these models were not scalable to large datasets and were unable to represent conditioned distributions in high-dimensional contexts such as images. 
Neural network-based models allowed for conditioning in high-dimensional variables such as images~\cite{levine2016end, james2018task} or text~\cite{lynch2020grounding, jang2022bc}, but they were typically trained as unimodal models. These types of models are at odds with the nature of the collected demonstrations. 
The inability of these models to capture the inherent diversity and multiple modes in the data led researchers to limit themselves to smaller~\cite{khansari2011learning} or highly curated datasets to ensure unimodality and thus simplify the modeling process.



\begin{figure}[t]
    \centering
      \begin{minipage}[c]{0.5\textwidth}
    	\includegraphics[width=.99\textwidth]{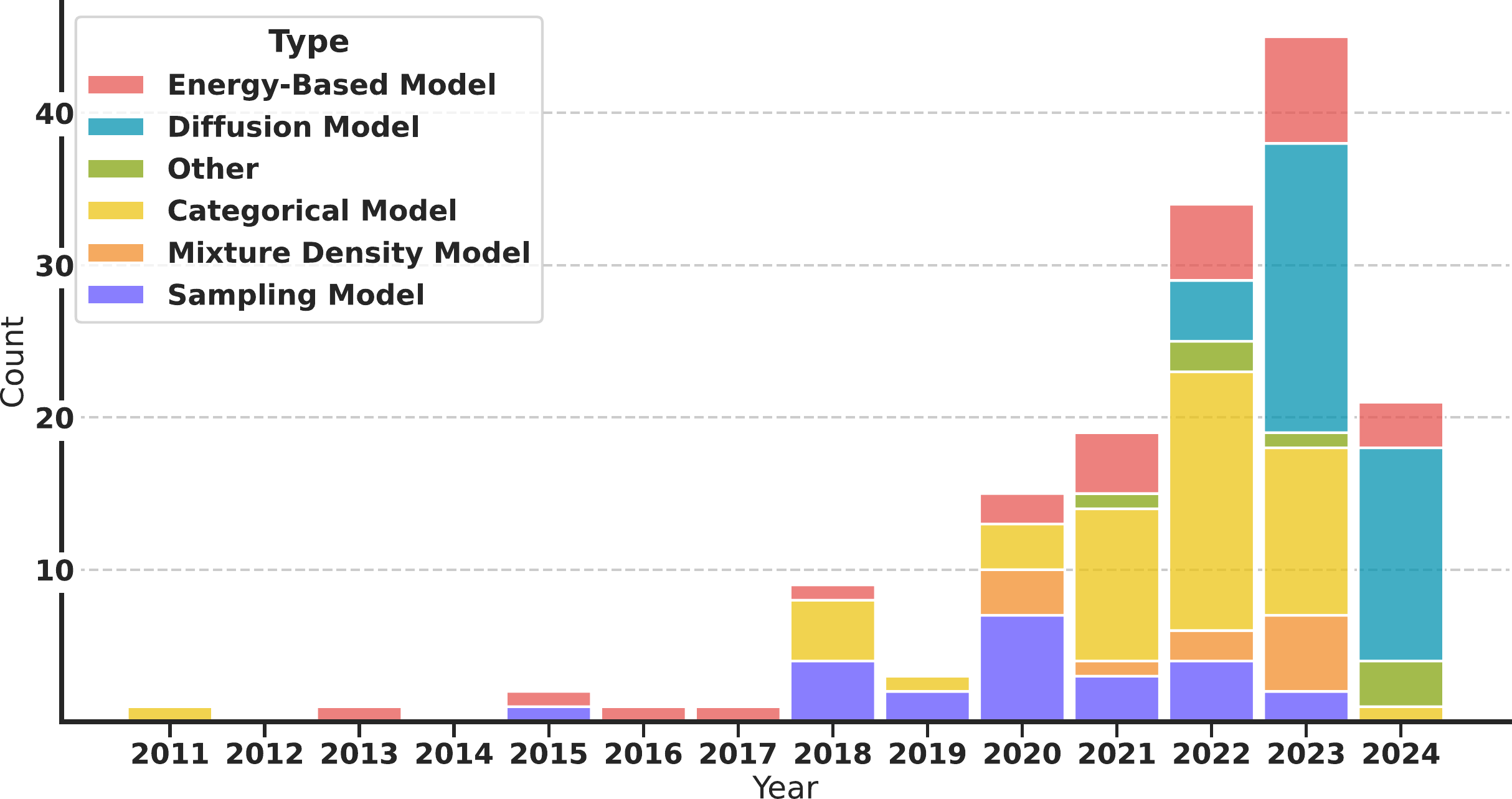}
    \end{minipage}\hfill
    \caption{Selected publications for this survey per year. Different colors indicate different types of \gls{dgm}. We categorize \gls{dgm} into five classes.}
    \label{fig:sec1_along_time_graph}
\end{figure}

Recent successes of \textbf{\gls{dgm}} in image~\cite{ho2020denoising} and text generation~\cite{devlin2018bert} have demonstrated their ability to capture highly multimodal data distributions.
In recent years, these expressive models have garnered attention in the field of robotics for Imitation Learning applications (see \Cref{fig:sec1_along_time_graph}).
For example, \gls{dm}~\cite{song2020score, ho2020denoising} have been effectively used to learn high-dimensional trajectory distributions \cite{janner2022planning, chi2023diffusion, reuss2023goal};
Language and image-based policies have been developed using GPT-style models representing categorical distributions in the action space
\cite{brohan2022rt}; 
and \gls{vae}~\cite{kingma2013auto} were applied to generate 6-DoF grasping poses for arbitrary objects~\cite{mousavian20196}.

This article presents a unified and comprehensive review of the various approaches explored by the robotics community to learn \gls{dgm} from demonstrations to capture the inherent multimodality of the data.
While some of these models are borrowed from other areas of machine learning, such as \gls{dm}, we also highlight approaches that have been particularly influential in representing action distributions in robotics, such as \textit{Action Value Maps} \cite{wu2020spatial, zeng2017robotic, ha2022flingbot}.

The survey focuses mainly on approaches that consider \textbf{offline data}, i.e., no additional data collected online or interactively, and \textbf{offline supervision}, i.e., no additional supervision other than expert actions.
Although learning \gls{dgm} from offline datasets has been widely studied in various fields from vision to text generation, there are inherent challenges in robotics that require careful design choices.
To motivate the specific design choices for robotics applications, in \Cref{sec:challenges_in_robotics}, we present the fundamental challenges of learning policies from demonstrations in robotics.

We divide the survey into six sections (see \Cref{fig:main}):
\\
In \textbf{\Cref{sec:problem_formulation}},
we formalize the problem and provide the nomenclature that we will use throughout the survey.
\\
In \textbf{\Cref{sec:type_of_dgm}}, we introduce the most commonly used \gls{dgm} in robotics, present their inherent properties, briefly list various works that have applied these methods to robotics, and present the training and sampling algorithms for each model.
\\
In \textbf{\Cref{sec:applications}}, we present the different types of applications in which deep generative models have been applied highlighting the type of data that the models generate and the type of conditioning variables that are considered.
\\
In \textbf{\Cref{sec:generalization}}, we present a range of design and algorithmic inductive biases to improve the generalization out of the data distribution of the learned models. 
How can we guarantee the generation of useful actions given as context observations that were not in the demonstrations?
Among the options we present are the modular composition of generative models, the extraction of informative features from the observations, and the exploitation of symmetries between the observations and the actions.
\\
Finally, in \textbf{\Cref{sec:research_challenges}}, we highlight the current research challenges in the field and suggest future directions.

\subsection{Challenges in Learning from Offline Demonstrations}
\label{sec:challenges_in_robotics}

Learning robot policies from offline demonstrations presents several challenges. While many of these challenges (e.g., multiple modes in the demonstrations) are shared with other research areas, such as image generation or text generation, there are robotics-specific challenges that we should consider.  Below, we present the main challenges in learning robot policies from offline data. 
\\
\textbf{Demonstration Diversity.}
One of the main challenges is the inherent variability within the demonstrations themselves~\cite{jia2024towards}. 
Different demonstrators may have different skill levels, preferences, and strategies for accomplishing the same task, resulting in a wide range of approaches encapsulated in the dataset.
Unimodal distributions lack the expressiveness to capture this variability in the demonstrations, resulting in poor performance.
\gls{dgm} are a promising approach to address this challenge. Being able to capture complex multimodal distributions, these models can learn to represent the different strategies and behaviors exhibited in the demonstrations.
\\
\textbf{Heterogeneous Action and State Spaces.}
Unlike computer vision, where the data space is well defined, in robotics, there is no a single state-action space. 
Robot actions can range from torque commands, to desired target positions or desired trajectories. In addition, robot behavior can be modeled in both the robot's configuration space and the task space.
This variability leads to heterogeneous datasets and heterogeneous solutions for learning robot policies.
\\
\textbf{Partially Observable Demonstrations.}
When a human performs a demonstration, his actions are not based solely on observable elements; they are driven by internal states influenced by the demonstrator's knowledge of the task and a history of observations. 
In addition, humans can incorporate information from the environment that may not be readily available or observable by a robot's sensors, such as peripheral details captured by human vision but missed by the robot's cameras. 
This mismatch often results in demonstrations that only partially represent the context of the task, leading to ambiguities in the policies learned by the robot.
The issue of partial observability has been studied extensively in the literature~\cite{spaan2012partially}.
A common practical approach is to encode the history of observations as contexts rather than a single observation, allowing the model to extract internal states that could reduce the ambiguity~\cite{mandlekar2022matters}.
\\
\textbf{Temporal Dependencies and Long-Horizon Planning.}
Robotic tasks often involve sequential decision-making, where actions are interrelated over time. This sequential nature can result in compounding errors that lead the robot into situations not encountered in the training demonstrations.
This problem has been addressed in several ways.
Some works propose learning short-horizon skills that can then be concatenated with a high-level planner.
In another direction, a number of works~\cite{janner2022planning, zhao2023learning} propose learning policies that generate trajectories of actions rather than single-step actions, thus reducing the sequentially compounded errors.
In addition, other options are to inject noise while generating the demonstrations~\cite{laskey2017dart} or to interactively grow the dataset~\cite{ross2011reduction}.
\\
\textbf{Mismatch between training and evaluation objectives.}
Learning from offline demonstrations is typically framed as a density estimation problem. The learned model is trained to produce samples that resemble the training dataset. 
However, the learned models are used to solve a given task, where the metric to be maximized is the task success rate.
This mismatch between the training objective and the evaluation objective can lead to poor performance when the robot is used to solve a particular task.
One possible direction to address this problem is to combine a behavioral cloning phase with a posterior reinforcement learning fine-tuning~\cite{kober2010imitation2}. 
\\
\textbf{Distribution Shifts and Generalization.} 
A fundamental challenge in learning from offline demonstrations is the distribution shift between the demonstration data and the real-world scenarios in which the learned policies are deployed. 
Demonstrations are typically collected in controlled environments or specific contexts, but the robot must operate in potentially novel situations not covered by the demonstrations. 
This mismatch can lead to generalization failures and performance degradation when the learned policies are applied outside the scope of the training data.
Addressing this challenge requires techniques that can extrapolate from the given demonstrations and adapt to new, unseen environments. We dedicate \Cref{sec:generalization} to explore different approaches to improve generalization in robotics applications.

\begin{figure*}[h]
    \centering
      \begin{minipage}[c]{0.99\textwidth}
    	\includegraphics[width=.99\textwidth]{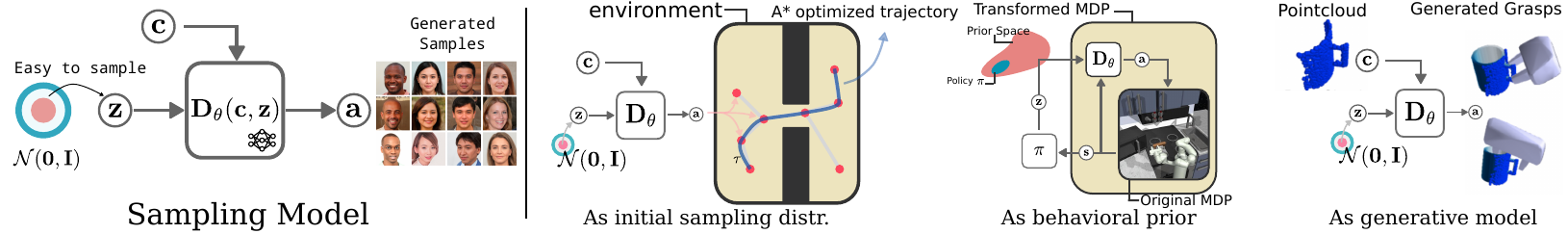}
    \end{minipage}
    \caption{Left: A visual representation of Sampling Models. Given a latent sample $\vz$, usually sampled from a normal distribution, Sampling Models generate an action sample through a learned decoder $\va = \mD_{\vtheta}(\vz,\vc)$. Right: A representation of common applications for Sampling Models: as sampling distribution~\cite{ichter2018learning}, as behavior prior~\cite{singh2020parrot} and, as generative model~\cite{mousavian20196}.}
    \label{fig:sec3_sampling_model}
    \vspace{-.5cm}
\end{figure*}

\subsection{Related Surveys}

The field of \gls{lfd} has a long history that has been explored in several surveys.
\\
Before deep learning-based approaches became standard, several surveys \cite{billard2008survey, argall2009survey, chernova2014robot, billard2016learning} explored the basic problem of Imitation Learning.
These surveys address questions such as \textit{How should we acquire data?}, \textit{What model should we learn?}, or \textit{How should we learn a policy?}.
\\
More recent works~\cite{hussein2017imitation, osa2018algorithmic, ravichandar2020recent} updated the reviews to the new state of the art where deep learning-based models were beginning to be integrated into \gls{lfd} problems. In particular, \cite{osa2018algorithmic} presented an algorithmic perspective on Imitation Learning, allowing the comparison of different algorithms from an information-theoretic point of view.

The current stage of the robot learning community, with the increasing availability of large-scale robot demonstrations both in simulation and in the real world, the growing importance of imitation-based approaches, and the increasing availability of cheap robot hardware, makes it timely to provide a survey covering the research of the last years and focusing on the challenges the field is currently facing (multimodality, generalization, heterogeneous datasets \dots).

Recently, a few surveys~\cite{firoozi2023foundation, hu2023toward} have explored the problem of learning foundational models for robotics, which mainly focused on integrating Internet-scale vision and language foundation models into robotics problems.
Despite the potential of applying vision-language foundational models to robotics problems, our survey focuses on a different problem.
The interest in this survey is in exploring approaches for learning policies directly from embodied robotics data (in part, due to the growing availability of large datasets~\cite{padalkar2023open, eppner2021acronym}), rather than adapting vision-language models to robotics.

\section{Problem Formulation}
\label{sec:problem_formulation}

The primary goal of \gls{bc} is to learn a conditioned probability density model (generative model) $\rho_{\vtheta}(\va|\vc)$, that accurately captures the underlying probability distribution of the data, denoted as $\rho_{\mathcal{\gD}}(\va|\vc)$, where $\va$ is the data variable we want to generate and $\vc$ is the conditioning variable.
The central idea is to ensure that the samples generated by the model $\va\sim\rho_{\vtheta}(\va|\vc)$ are indistinguishable from the real data samples $\va\sim\rho_{\mathcal{\gD}}(\va|\vc)$.

In the context of decision-making and control, $\va$ represents the action, which range from end-effector poses \cite{urain2022se3dif}, displacements \cite{shafiullah2022behavior}, trajectories \cite{janner2022planning}, desired scene arrangement \cite{liu2022structformer}, to robot configurations \cite{lembono2020generative}.
The conditioning variable $\vc :(\vo, \vg)$ is usually decoupled between $\vo$ the observations of the scene and $\vg$ the goal definition.
Observations may include visual data~\cite{zeng2021transporter}, 3D spatial data~\cite{mo2021where2act}, or robot proprioception, providing information about the state of the environment.
Depending on the task, it is also common to provide a history of the last $t$ observations rather than a single-step observation.
The goal variable $\vg$ defines the desired behavior or task that the robot should accomplish.
This goal can be specified in a variety of ways, including language commands~\cite{ha2023scalingup}, desired goal states~\cite{ames2022ikflow}, or goal images~\cite{mandlekar2020learning}; each provides a different approach to directing the robot's actions toward achieving specific outcomes.

To learn the model $\rho_{\vtheta}(\va|\vc)$, we operate under the assumption that the true data distribution $\rho_{\mathcal{\gD}}(\va|\vc)$ is unknown and that we only have access to a finite set of samples drawn from that distribution. 
These samples form a dataset $\gD:\{\va_n, \vc_n\}_{n=1}^{N}$ where $N$ is the number of samples.
The task of learning the generative model is then formulated as an optimization problem, where the objective is to minimize the divergence between the learned distribution $\rho_{\vtheta}(\va|\vc)$ and the true data distribution $\rho_{\gD}(\va|\vc)$
\begin{align}
    \vtheta^* = \argmin_{\vtheta} \E_{\va,\vc\sim\gD}\left[ \sD (\rho_{\gD}(\va|\vc),\rho_{\vtheta}(\va|\vc)) \right],
    \label{eq:div_min}
\end{align}
where $\sD$ is the divergence distance.
Despite the general representation in \eqref{eq:div_min}, the training algorithm is modified depending on the selected model $\rho_{\vtheta}(\va|\vc)$ (Gaussian, \gls{ebm}~\cite{lecun2006tutorial, song2021train}, \gls{dm}~\cite{song2019generative, ho2020denoising}).

\section{Density Estimation Models}
\label{sec:type_of_dgm}

The central idea of this survey is to present in a unified way the different types of models that have been used in robotics to properly capture the multimodality in the demonstrations.
Thus, this survey does not include works that have used unimodal models to represent the policies, and focuses on models that are able to generate samples from multimodal distributions.
We categorize these models into five groups:
\\
\textbf{Sampling Models.} Given a noise sample, these models generate the action directly. They tend to have very fast inference times. \gls{dgm} like \gls{vae}, \gls{gan}, or \gls{nf} fall into this category.
\\
\textbf{Energy-based Models.} Given an action candidate as input, \gls{ebm} returns a scalar value representing the energy of that action candidate. Sampling from a \gls{ebm} usually requires \gls{mcmc} strategies. We also consider as \gls{ebm}, models that define the energy as the distance between feature descriptors~\cite{simeonov2022neural}.
\\
\textbf{Diffusion Models.} \gls{dm} are a type of generative model that learns to generate data by reversing a gradual corruption process. These types of models are able to generate high quality samples due to the iterative denoising process.
\\
\textbf{Categorical Models.} Given a context variable, categorical models represent the action distribution as a discrete distribution of  $k$ bins. We group both GPT-inspired action models~\cite{brohan2022rt} and action value maps~\cite{zeng2021transporter} into this category. Note that despite the categorical distributions represent both types of models, action value maps directly inpaint the categorical distribution in the visual observations. In contrast, in GPT-inspired models, observations and action distributions are represented in different spaces.
\\
\textbf{Mixture Density Models.} Given a context variable, \gls{mdm} returns the parameters of a mixture density function representing the action distribution. Common choices are models that return the means, standard deviations, and weights of a \gls{gmm} or a mixture of logistic distributions.

The classification presented is not rigid or definitive. For example, Normalizing Flows~\cite{rezende2015variational} operates as a sampling model in the generation, but it also facilitates the calculation of the likelihood of a sample in a manner akin to \gls{ebm}. Furthermore, we cluster inside Categorical models to both GPT-style autoregressive models~\cite{brohan2022rt, brohan2023rt} and Action Value Maps~\cite{zeng2018learning, zeng2020tossingbot}. 
While both models express the distribution via a categorical distribution, they diverge conceptually. 

In the following, we present in five distinct subsection each model type, its inherent properties and the type of problems in which it has been applied.

\subsection{Sampling Models}
We call sampling models to the set of deep generative models that allow explicit sample generation.
Given a context variable $\vc\in\RR^c$ and a latent variable $\vz\in\RR^z$, the network decodes the latent variable into a sample $\va = \mD_{\vtheta}(\vz, \vc)$. 
To generate an action sample from our model $\va \sim \rho_{\vtheta}(\va|\vc)$, we first sample a latent variable from an easy-to-sample from distribution $\vz \sim \gN(\vzero, \mI)$ (e.g., a normal distribution) and decode it into an action $\va = \mD_{\vtheta}(\vz, \vc)$ (see \Cref{fig:sec3_sampling_model}).
There are several generative models that fall into this category: \gls{gan}~\cite{goodfellow2014generative}, \gls{vae}~\cite{kingma2013auto}, or \gls{nf}~\cite{rezende2015variational}.

\subsubsection{\textbf{Main applications}}
In the field of robotics, these types of models have been used in several contexts and applications (see \Cref{fig:sec3_sampling_model}).
\\
\textbf{As an Initial Sampling Distribution.}
Due to their fast sampling time, they have been used as initial sampling distributions for motion planning and optimization problems~\cite{ichter2018learning, mohammadi2018path, ortiz2022structured, lai2021plannerflows, lai2022parallelised}. 
In \cite{ichter2018learning}, conditioned \gls{vae} were used to sample initial collision-free guided states for sampling-based motion planning problems~\cite{lavalle1998rapidly, kavraki1996probabilistic}.
In \cite{ortiz2022structured}, \gls{gan} were used to generate initial states for long-horizon tasks and motion planning problems. The output of the \gls{gan} was later optimized to satisfy a set of constraints.
\\
\textbf{As Exploration Guiding Models.}
A common problem in \gls{rl} is the exploration. Given the large state-action space, deciding which regions are meaningful to explore is usually a hard problem. To guide this exploration, several works~\cite{van2016stable, singh2020parrot, pertsch2020accelerating, lynch2020learning, mandlekar2020iris} have explored learning a sampling model that encodes all the possible behaviors in a dataset.
This model can be used integrated into a \gls{rl} problem, by running a policy in the latent space. Given that the model will generate solutions from the dataset, the policy learns to search in the latent space to maximize a given reward.
\\
\textbf{As Explicit Sampling Models.}
The most straightforward application is to use the model as a generative model.
In this context, sampling models have been used to generate grasp poses~\cite{mousavian20196, yan2019learning}, inverse kinematic
solutions~\cite{lembono2020generative, ames2022ikflow}, or directly sample actions in a policy~\cite{urain2020imitationflows, rana2020euclideanizing}.

\subsubsection{\textbf{Training Sampling Model}}

\gls{gan}, \gls{vae}, and \gls{nf} share the same sampling process.
However, each model is trained using a different algorithm.
In the following, we briefly present the training pipelines for the three models.
\\
\textbf{Variational Autoencoders.}
The \gls{vae} model, introduced in \cite{kingma2013auto}, consists of two networks: an encoder and a decoder. 
Given an action $\va$, the encoder maps it to the parameters of a latent normal distribution $\vmu_z, \vsigma_z = \gE_{\vpsi}(\va)$.
Given a sample from the latent space $\vz\sim \gN(\vmu_z, \vsigma_z\mI)$, the decoder maps the latent variable to the action space $\hat{\va} = \mD_{\vtheta}(\vz,\vc)$, conditioned on the context variable $\vc$.

The training loss consists of two parts: a reconstruction loss and a KL divergence. Given a dataset $\gD: \{ \va_n, \vc_n \}_{n=1}^N$, the \gls{vae} loss is given by
\begin{align}
    \gL(\vtheta, \vpsi) =& \E_{\va, \vc \sim \gD} [ \sD_{\text{KL}}(\rho(\vz|\va), \rho(\vz)) \\ & + \E_{\vz\sim\rho(\vz|\va)}[ || \mD_{\vtheta}(\vz, \vc) - \va ||_2^2 ] ]\nonumber
\end{align}
where $\rho(\vz|\va) = \gN(\vz| \gE(\va))$ is a Gaussian whose parameters are the encoder outputs.
$\rho(\vz)=\gN(\vzero, \mI)$ is a Gaussian around zero.
While the KL divergence term encourages the encoder to generate distributions close to $\rho(\vz)$, the reconstruction loss aims to decode a latent sample to look as similar as possible to the input $\va$.
\\
\textbf{Generative Adversarial Networks.}
Unlike \gls{vae}, \gls{gan} \cite{goodfellow2014generative} suggests having a discriminator $p = C_{\vpsi}(\va,\vc)$ instead of an encoder.
Given a sample generated by the model $\va \sim \rho_{\vtheta}(\va|\vc)$, the discriminator is trained to discriminate between samples generated by our model and samples coming from the dataset, while the generator is trained to make the generated samples as similar as possible to the dataset.

Given a dataset $\gD: \{ \va_n, \vc_n \}_{n=1}^N$, the \gls{gan} objective is represented by the binary cross-entropy loss
\begin{align}
     \gJ(\vtheta, \vpsi) =& \E_{\va,\vc \sim \gD}[ \log C_{\vpsi}(\va,\vc) + \\ & \E_{\vz \sim \gN(\vzero,\mI)}\left[ \log (1 - C_{\vpsi}(\mD_{\vtheta}(\vz,\vc)),\vc) \right]] \nonumber.
\end{align}
Then, the optimization problem is solved by a minimization-maximization problem, where we aim to minimize the objective with respect to $\vpsi$ (the discriminator) and maximize it with respect to $\vtheta$ (the generator).
The discriminator aims to discriminate between real data samples and the fake samples produced by the generator, while the generator aims to produce samples that are indistinguishable from real data to the discriminator.
\\
\textbf{Normalizing Flows.}
The generator $\mD_{\vtheta}$ in \gls{nf} is different from those in \gls{vae} or \gls{gan}. 
While in \gls{gan} or \gls{vae} it is represented by an arbitrary network, in \gls{nf} we are required to have an invertible network~\cite{rezende2015variational, papamakarios2019normalizing,chen2018neural} as generator.

Since the generator $\mD_{\vtheta}$ is invertible, \gls{nf} allows the exact calculation of the likelihood \cite{rezende2015variational}
\begin{align}
    \log \rho_{\vtheta}(\va|\vc) = \log \rho_z \left( \mD_{\vtheta}^{-1}(\va,\vc)\right) + \log|\det \mJ_{\mD_{\vtheta}}(\va,\vc)|,
    \label{ch2:eq_nflows}
\end{align}
where $\rho_z = \gN(\vzero, \mI)$ is the latent space normal distribution and $\mJ_{\mD_{\vtheta}}$ is the Jacobian of the decoder.

Then, given a dataset $\gD: \{ \va_n, \vc_n \}_{n=1}^N$, \gls{nf} are trained by minimizing the negative log-likelihood
\begin{align}
    \gL(\vtheta) &= -\E_{\va, \vc \sim\gD}\left[ \log  \rho_{\vtheta}(\va|\vc) \right].
\end{align}
Note that unlike \gls{vae} and \gls{gan}, \gls{nf} does not require to training an additional model. Additionally, since the generator is invertible, we can compute the likelihood of a sample in our model, similar to \gls{ebm}.


\subsection{Energy-Based Models.}
\begin{figure*}[t]
    \centering
      \begin{minipage}[c]{0.99\textwidth}
    	\includegraphics[width=.99\textwidth]{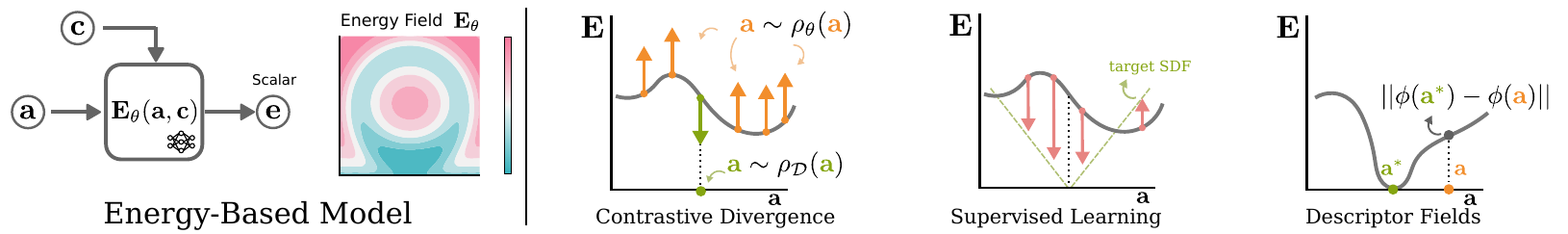}
    \end{minipage}
    \caption{Left: A visual representation of an \gls{ebm}. Given as input an action variable $\va$, \gls{ebm} output the unnormalized log probability of the input action $e = \mE_{\vtheta}(\va,\vc)$. Right: A visual representation of the different strategies to train or represent an \gls{ebm}: Contrastive Divergence~\cite{finn2016deep}, Supervised Learning~\cite{weng2023neural} and, Neural Descriptor Fields~\cite{simeonov2022neural}.}
    \label{fig:sec3_ebm}
    \vspace{-.1cm}
\end{figure*}

We call \gls{ebm} to the set of deep generative models that, given an action $\va$, output a scalar value $\ve \in \RR$, $\ve = E_{\vtheta}(\va, \vc)$, where $\vc$ denotes the conditioning context variable (\Cref{fig:sec3_ebm}).
In \gls{ebm}, the probability density model $\rho_{\vtheta}(\va|\vc)$ is represented by a Boltzmann distribution
\begin{align}
    \rho_{\vtheta}(\va|\vc) \propto \exp \left( -E_{\vtheta}(\va,\vc) \right),
\end{align}
where $E_{\vtheta}(\va,\vc)$ is the energy of the distribution, i.e., the unnormalized log-likelihood.

Sampling from an \gls{ebm} $\va \sim \rho_{\vtheta}(\va|\vc)$ is not direct due to the implicit nature of the model. \gls{ebm} define a probability distribution over the data by an energy function, and sampling requires methods like \gls{mcmc} to approximate the distribution.
A common sampling algorithm is \textit{Langevin Monte Carlo}. Given an initial sample $\va_0 \sim\rho_0(\va_0)$ generated from a simple prior distribution, the samples are generated by iteratively updating the sample by
\begin{align}
    \va_{k+1} = \va_{k} - \frac{\epsilon}{2}\nabla_{\va}E_{\vtheta}(\va_k,\vc) + \sqrt{\epsilon}\gN(\vzero, \mI),
\end{align}
where $\epsilon>0$ is a small constant.
This process can be computationally intensive and slower than direct sampling methods used in models such as \gls{vae} or \gls{gan}.
Alternatively, given the implicit nature of the \gls{ebm}, some works \cite{urain2021composable2, florence2022implicit} search for the most likely sample by solving an optimization problem
\begin{align}
    \va^* = \arg \min_{\va} E_{\vtheta}(\va,\vc).
\end{align}

Because of their implicit nature, \gls{ebm} contain several interesting properties.
As explored in \cite{du2019compositional, urain2021composable2, gkanatsios2023energy}, \gls{ebm} allow a \textit{modular composition} of different \gls{ebm}.
This modular approach allows separate \gls{ebm} to be trained to represent different behaviors or aspects of the data, and then these models can be combined. 
The result is a composite model in which the variable $\va$ has a high probability under all the component models, effectively integrating different features or patterns captured by each individual \gls{ebm}.

Another interesting property is the \textit{energy mapping}~\cite{urain2022se3dif, gkanatsios2023energy}.
\gls{ebm}, due to its implicit nature, allows to generate samples in a space different from their training space. This is especially useful in robotics. For example, \gls{ebm} trained in the task space $\mE_{\vtheta}(\vx|\vc)$ can effectively guide the selection of robot joint configurations $\vq$, if we have access to the forward kinematics mapping $\vx = \phi_{\text{FK}}(\vq)$.
By composing the map and the energy $\mE_{\vtheta}(\vq|\vc) = \mE_{\vtheta}(\phi_{\text{FK}}(\vq)|\vc)$, we can represent an \gls{ebm} in the configuration space that sets low energy to those configurations that lead to low energy in the task space.


\subsubsection{\textbf{Main Application}}
The main applications of \gls{ebm} in robotics range from cost/reward functions for sequential decision making problems to direct generative models.
\\
\textbf{As a cost/reward function.}
Learning \gls{ebm} to represent cost or reward functions has been widely studied in \gls{ioc}~\cite{kalakrishnan2013learning, finn2016guided} or \gls{irl}~\cite{ziebart2008maximum, fu2017learning}.
Some training algorithms such as \gls{cd} loss, require the generation of samples from the learned \gls{ebm} during the training process. Different \gls{irl} and \gls{ioc} methods propose different approaches to sample from the learned \gls{ebm}.
\cite{finn2016guided} proposes solving a maximum entropy trajectory optimization to generate the samples, \cite{du2019implicit} proposes generating the samples using Langevin dynamics, while \cite{ziebart2008maximum, fu2017learning}, propose generating the samples from a policy trained with \gls{rl}, given the learned \gls{ebm} is the reward function.
\\
\textbf{As a generative model.}
Besides \gls{irl} and \gls{ioc} approaches, which focus on sequential decision-making problems, several works have explored learning \gls{ebm} for direct action generation.
In \cite{weng2023neural}, an \gls{ebm} is learned to generate grasping poses for arbitrary objects. Unlike most of the approaches to learning \gls{ebm}s, the model is learned to fit a 6-DoF \gls{sdf}.
Similarly, \cite{gervet2023act3d} generates end-effector poses. However, their work focuses on a broader set of tasks beyond grasping, such as opening drawers or pressing buttons.
In \cite{florence2022implicit}, an \gls{ebm} is trained as a visuomotor policy. The authors claim that a \gls{ebm}-based policy captures the discontinuities in the data better than a deterministic or Gaussian policy. 
In \cite{Du2019ModelBP}, an \gls{ebm} is trained as a transition dynamics model.
In \cite{sodhi2022leo}, a cost function for state estimation is learned. The learned \gls{ebm} represents the joint probability of the state given an observation.

\subsubsection{\textbf{Training Energy-Based Model}}
One of the most popular algorithms for training \gls{ebm} is
\gls{cd}\cite{hinton2002training, du2019implicit}.
This method involves a contrastive game between negative samples, generated from a given distribution and positive samples, obtained from the dataset.

In contrast to \gls{cd}, it has been popular in robotics to train models with supervised learning losses such as occupancy loss or \gls{sdf}.
Despite not being common, given their popularity in robotics, we introduce them as additional approaches to fitting \gls{ebm} (see \Cref{fig:sec3_ebm}).
\\
\textbf{Contrastive Divergence.}
A common approach to learning density models is to minimize the negative log-likelihood of the data.
However, computing the log-likelihood requires access to the 
model's normalization constant, which is intractable in \gls{ebm}.
To adapt the negative log-likelihood loss to \gls{ebm}, \cite{hinton2002training} approximates the calculation by samples.
Given $\nabla_{\vtheta} \log Z_{\vtheta} = -\E_{\va \sim \rho_{\vtheta}(\va)} \left[ \nabla_{\vtheta}E_{\vtheta}(\va) \right]$, we can approximate the gradient of the negative log-likelihood
\begin{align}
    \nabla_{\vtheta}\gL(\vtheta) = \E_{\textcolor{greenhak}{\va}\sim \gD} \left[ \nabla_{\vtheta}E_{\vtheta}(\textcolor{greenhak}{\va}) \right] - \E_{\textcolor{orangehak}{\va}\sim \rho_{\vtheta}} \left[ \nabla_{\vtheta}E_{\vtheta}(\textcolor{orangehak}{\va}) \right].
    \label{eq:ch2_cd}
\end{align}
In \Cref{eq:ch2_cd}, the energy is pushed down for the samples in the dataset (positive samples) and pushed up for the rest of the samples that are not part of the dataset.
On each iteration, we sample a set of points from the current energy model $\va \sim \rho_{\vtheta}(\va)$ (negative samples) to evaluate the model.
However, if the dimension of $\va$ is large, it may be difficult to sample $\va \sim \rho_{\vtheta}(\va)$ properly. To mitigate this difficulty, techniques such as using the gradient of the energy function ~\cite{du2019implicit} or minimizing the KL divergence between samples ~\cite{du2020improved} are common approaches.
\begin{figure*}[t]
    \centering
      \begin{minipage}[c]{0.99\textwidth}
    	\includegraphics[width=.99\textwidth]{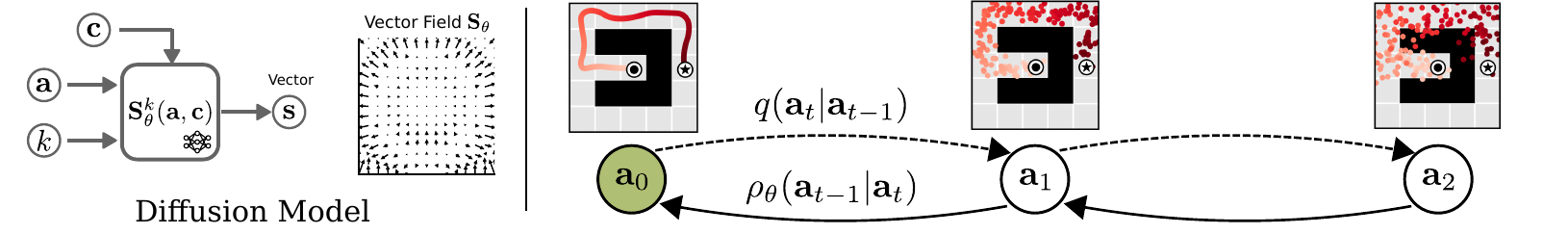}
    \end{minipage}
    \caption{Left: A visual representation of a \gls{dm}. Given an action $\va$ and a scalar $k$ informing on the diffusion step, the model $\vs=\mS_{\vtheta}(\va,\vc,k)$ outputs a vector $\vs$ conditioned on $\vc$. The output $\vs$ is related to the score of a distribution $\rho(\va_k)$. Right: A visualization of the denoising process \cite{janner2022planning}.}
    \label{fig:sec3_diffusion}
    \vspace{-.1cm}
\end{figure*}
\\
\textbf{Supervised Learning.}
A few works in robotics~\cite{weng2023neural, jiang2021synergies} have explored training \gls{sdf}, occupancy fields, or binary classifiers to represent scalar fields.
Although not explicitly trained as a generative model, the learned model is architecturally equivalent to an \gls{ebm}.
The model takes an action variable $\va$ as input and outputs a scalar value $\ve$ that informs about the quality of the sample. 
After the training, this model can be used to generate samples. For example, in \cite{weng2023neural} a \gls{sdf} model is trained to generate 6 DoF grasp poses, while in \cite{jiang2021synergies} an occupancy network is trained to generate grasp poses. 
\\
\textbf{Neural Descriptor Fields.}
A set of works~\cite{simeonov2022neural, simeonov2023se} represent an \gls{ebm} as the Euclidean distance to a target action $\va^*$ in a learned latent space $E(\va|\va^*) = || \vphi_{\vtheta}(\va^*) - \vphi_{\vtheta}(\va) ||$, where $\vz = \vphi_{\vtheta}(\va)$, maps an action to a latent vector $\vz\in\RR^d$ of dimension $d$.
In contrast to learning the \gls{ebm} directly, these methods propose learning a feature encoder $\vphi_{\vtheta}$, which computes a latent vector for a given input $\va$.
In \cite{simeonov2022neural, simeonov2023se}, the feature encoder is trained by reconstructing the \gls{sdf} of an object. The feature encoder is conditioned on the pointcloud of the object $\vphi_{\vtheta}(\va,\vc)$, where $\vc$ is the pointcloud.
In \cite{shafiullah2022clip}, the CLIP~\cite{radford2021learning} features are used to learn the feature encoder.

\subsection{Diffusion Models}
\label{sec:models_dm}
\gls{dm}~\cite{sohl2015deep, ho2020denoising} frame the data generation process as an iterative denoising process.
Given a prior sampling distribution $\va_N \sim \rho(\va_N)$, typically a Gaussian distribution, an iterative denoising process $\rho_{\vtheta}(\va_{k-1}|\va_{k})$ moves the noisy samples from the prior to the data distribution
\begin{align}
    \rho(\va_0) = \int \rho(\va_N) \prod_{k=1}^{N} \rho_{\vtheta}(\va_{k-1}|\va_k) \d \va_{1:N},
\end{align}
where $ \rho(\va_0) \equiv \rho_{\gD}(\va_0)$ is equivalent to the data distribution. The denoising process $\rho_{\vtheta}(\va_{k-1}|\va_k)$ is the inverse of a forward diffusion process $q(\va_{k+1}|\va_k)$ that gradually adds noise to the dataset samples. 

In practice, \gls{dm} are closely related to \gls{ebm}, where the denoising prediction estimates the gradient field of an energy function ~\cite{vincent2011connection, liu2022compositional,du2023reduce}.
Given as input an action $\va$, outputs a vector $\vs$, $\vs = \mS_{\vtheta}(\va, \vc, k)$, where $\vc$ denotes the context variable and $k$ is a scalar value informing about the diffusion step, each step of the diffusion process can be seen as a step of Langevin dynamics sampling using an \gls{ebm}.
Due to the iterative sampling process, \gls{dm} have slower inference times compared to other \gls{dgm}. Recent research, such as Consistency Policies~\cite{prasad2024consistency}, explores how to make \gls{dm} sampling faster. 



\gls{dm} have become particularly popular in the last year because of several important properties for generative modeling. \gls{dm} are capable of representing high-dimensional continuous space distributions and have a stable training pipeline.
Due to its connection to \gls{ebm}, \gls{dm} implicitly parameterizes the actions, allowing the composition of diffusion models.
Several works have explored the \textit{modular composition} of \gls{dm} with additional objectives~\cite{janner2022planning, urain2022se3dif, ajay2022conditional, carvalho2023motion} or other \gls{dm}~\cite{mishra2023generative, yang2023compositional}. 
Although not applied to robotics, composition is particularly popular in image generation. Classifier Guidance~\cite{dhariwal2021diffusion} proposes to combine the output of an unconditional \gls{dm} with the gradient of a classifier in the generation process. Classifier-free guidance~\cite{ho2022classifier} proposes instead to combine an unconditional \gls{dm} with a conditioned \gls{dm}.
In \cite{liu2022compositional, huang2023composer}, several conditioned \gls{dm} were combined for modular generation. 

\subsubsection{\textbf{Main Applications}}
Due to their high expressiveness and flexibility, \gls{dm} have been widely integrated into many robotics tasks. We decouple the applications into three main clusters: papers that aim to directly learn robot trajectories, papers that explore the modularity and composability of \gls{dm}, and papers that generate other types of variables, beyond trajectories.

\textbf{Trajectory Generation.}
The generation of robot trajectories is an essential element for solving any robot task.
Traditionally, these trajectory generators have been represented with simpler models, such as policies that generate the trajectories autoregressively, or structured models, such as task and motion planning algorithms. 
However, the expressiveness of \gls{dm} has allowed the robotics community to directly generate the trajectories without the need for these models.
In \cite{janner2022planning, ajay2022conditional}, \gls{ddpm} was used to learn trajectory generators from demonstrations.
The generation of the robot trajectories is similar to a receding horizon control loop~\cite{kwon2005receding}, which allows reactive generation.
In \cite{reuss2023goal, chi2023diffusion} the trajectory \gls{dm} was introduced conditioned on visual input and in \cite{ha2023scalingup, reuss2023multimodal} it was conditioned on both language and vision.
\cite{carvalho2023motion, huang2023diffusion} explored the integration of \gls{dm} for motion planning problems, both for collision-free trajectory generation and navigation tasks.

\textbf{Generation beyond trajectories.}
Besides trajectories, \gls{dm} have been used to generate several types of data for robotics tasks.
Several works~\cite{urain2022se3dif, simeonov2023shelving, mishra2023reorientdiff} have explored using \gls{dm} to generate \texttt{SE(3) poses}, both for generating robot grasp poses~\cite{urain2022se3dif} or object placement poses~\cite{simeonov2023shelving, mishra2023generative} for pick and place tasks.
Some works~\cite{liu2022structdiffusion, kapelyukh2022dall, yang2023compositional} have used \gls{dm} to generate \texttt{scene arrangements}. This information is used to define a high-level goal plan for a motion planner to rearrange a scene.
Some works~\cite{du2023video, du2023learning, ajay2023compositional} have explored learning \texttt{video} \gls{dm}, which are used to have a general representation plan of the desired behavior. Then an inverse dynamics model generates actions in the robot to match the video behavior.
While \cite{higuera2023learning}, \gls{dm} were used to generate realistic tactile images from images; in \cite{scheikl2024movement} movement primitive weights are generated.

\textbf{Modular Composition.}
Several of the papers cited above have explored modular composition of \gls{dm} in various forms.
In \cite{janner2022planning, ajay2022conditional}, the learned \gls{dm} is combined with a reward function to condition the generation towards high-reward regions.
In \cite{carvalho2023motion}, a trajectory \gls{dm} is combined with the \gls{sdf} of the scene to perform collision free trajectory generation.
In \cite{urain2022se3dif}, a generative grasping model was integrated into a motion planning problem to generate trajectories for pick and place tasks.
\cite{yang2023compositional} composes multiple object relations to generate scene arrangements with local models.
In \cite{mishra2023reorientdiff}, the placement generative model is combined with a feasibility score to adapt to feasible placement poses.
In \cite{mishra2023generative}, instead, a set of \gls{dm} are temporally composed to solve long horizon planning tasks.

\subsubsection{\textbf{Training Diffusion Models}}

An interesting property of \gls{dm} is their stable training process. Unlike \gls{cd} for training \gls{ebm}, \gls{dm} are learned with stable target signals, which leads to more robust training, but can still be seen as a parameterization of an implicit energy landscape. Below we describe the training of the diffusion models and their connection to \gls{ebm}.

To train a diffusion model, we learn to model the inverse transition function $\rho_{\vtheta}(\va_{k-1}|\va_{k})$ at each time step $k$. This transition kernel is parameterized by a Gaussian distribution corresponding to the sampling process
\begin{equation}
    \va_{k-1} = B_k (\va_k - C_k \mS_{\vtheta}(\va_k, k) + D_k \mathbf{\xi}), \quad \mathbf{\xi} \sim \mathcal{N}(\mathbf{0}, \mathbf{I})
    \label{eqn:diffuse_langevin}
\end{equation}
where $B_k$, $C_k$, and $D_k$ are fixed (unlearned) constants in the diffusion process.  To learn the above transition distributions, we only need to learn and model the score function $\mS_{\vtheta}(\va, k)$. 

The score function $\mS_{\vtheta}(\va_k, k)$ corresponds to the gradient of an \gls{ebm} $\mS_{\vtheta}(\va, k) = \nabla_{\va}E_\theta(\va, k)$, where the \gls{ebm} models the noise convolved action distribution $p_k(\va) \;\propto\;  e^{-E_\theta(\va, k)}$, where
\begin{equation}
    p_k(\va) = \int_{\va^*} p(\va^*) \cdot \mathcal{N}(\va; \sqrt{1-\sigma_k^2} \va^*, \sigma^2_k \mathbf{I}).
\end{equation}
We can directly learn this \gls{ebm} $E_\theta(\va, k)$ implicitly by directly training the $\mS_{\vtheta}(\va, k)$ to denoise actions through denoising score matching ~\cite{vincent2011connection}. 
\begin{equation}
    \label{eqn:denoise}
      \mathcal{L}_{\text{MSE}}(\theta) = \| \mS_{\vtheta}(\va_k + \epsilon, k) - \epsilon \|^2, \quad \epsilon \sim \mathcal{N}(0, 1).
\end{equation}
In comparison to methods for training \gls{ebm}, this objective to learn the score is both faster and more stable.

By learning this score function  $\mS_{\vtheta}(\va_k, k)$ and corresponding implicit landscape $E_\theta(\va, k)$ in a diffusion model, each reverse transition kernel in the diffusion process in \Cref{eqn:diffuse_langevin} corresponds to Langevin sampling on a sequence of noise convolved \gls{ebm} $E_\theta(\va, k)$, where an added contraction term $B_k$ is used to transition between separate successive energy landscape. This implicit view of sampling in diffusion models allows us to combine multiple diffusion models together~\cite{du2023reduce}.





\subsection{Categorical Models}
We refer to Categorical Models as the set of generative models that, given a context variable $\vc$ as input, output the probability of K different categories, where K is finite. 
In practice, it is common for the network to output K logits.
Given the logits, a Softmax function converts the logits into probability values for each action.

\begin{figure}[t]
    \centering
      \begin{minipage}[c]{0.39\textwidth}
    	\includegraphics[width=.99\textwidth]{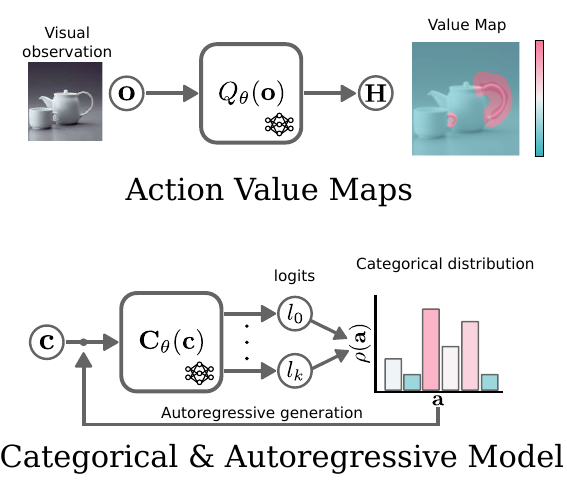}
    \end{minipage}
    \caption{A visual representation of Action Value Maps and Autoregressive models. Given a visual observation $\vo$ as input, Action Value Maps $\mH = Q_{\vtheta}(\vo)$ output an observation shape probability map $\mH$, where each pixel is a projection of a possible action. Autoregressive models output K}
    \label{fig:category_model}
    \vspace{-.5cm}
\end{figure}

In robotics, categorical models are mainly used to represent policies. There are two main types of architectures in which categorical distributions have been used: Action Value Maps and Autoregressive Models.
\\
\textbf{Action Value Maps}
\\
Action Value Maps are a set of generative models that, given a visual observation $\vo$ as input, output a value map $\mH$ of the same form as the input, $\mH = Q_{\vtheta}(\vo)$, similar to attention maps. The visual observation $\vo$ can be images~\cite{zeng2017robotic,goyal2023rvt}, point clouds~\cite{mo2021where2act}, or voxels~\cite{breyer2021volumetric,shridhar2023perceiver}. The generated value map $\mH$ assigns a probabilistic value to each distinct location or `pixel' within the visual observation $\vo$.

The core principle behind action value maps is the interpretation of each `pixel' location in the visual input as representative of a potential action that a robot could perform. Consider, for example, a robot picking task. 
In this scenario, each individual pixel in an image could represent a potential target point for moving the robot arm to perform a pick. The resulting value map is then analyzed as a categorical distribution encompassing the full range of potential actions.
Interestingly, the approach is easily extended to multiple primitives, where each pixel represents the geometrically grounded parameters of each primitive. 
For example, in \cite{zeng2018learning} two value maps are generated, one for the grasping primitive and another for the pushing primitive. Then the selected action is the one with the highest energy among all possible pixels.

A number of works~\cite{shridhar2022cliport, shridhar2023perceiver, goyal2023rvt} have extended action value maps to also be conditioned on language commands, defining the goal $\vg$ of the learned policy $Q_{\vtheta}(\vo, \vg)$.
\\
\textbf{Autoregressive Models}
\\
Autoregressive models are a set of generative models that generate long-horizon data by iteratively invoking the model while conditioning on past data. For example, given an action trajectory $\vtau = (\va_0, \va_1, \dots, \va_T)$, an autoregressive model represents the distribution over the trajectory
\begin{align}
    \rho(\vtau) = \rho(\va_0)\prod_{i=1}^{T} \rho(\va_i|\va_{0:i-1}),
\end{align}
as the product of the conditioned action distribution on the previous data. To generate a trajectory, we first sample an initial sample $\va_0 \sim \rho(\va_0)$ and then iteratively call the generative model conditioned on the previously generated actions.
This type of model has become particularly popular for language generation. Models such as BERT~\cite{devlin2018bert} or GPT-3~\cite{brown2020language} build on Transformer models~\cite{vaswani2017attention} to generate autoregressive long text data. Some generative models in image generation are also based on autoregressive generative models, where the pixels of the image are generated iteratively~\cite{van2016conditional}.

In robotics, it has been common to refer to autoregressive models as policies that use GPT-like structures to generate robot actions~\cite{ahn2022can, brohan2022rt, brohan2023rt, driess2023palm, lee2022multi, reed2022generalist, janner2021offline}. 
Several of these models represent the action distribution (i.e., policy) as a categorical distribution and generate action trajectories autoregressively, similar to language models. However, in certain cases, the action distribution is represented by \gls{dm}~\cite{team2023octo, jia2024mail} or by adapting the means of the categorical distribution~\cite{shafiullah2022behavior, cui2022play, lee2024behavior}.

\subsubsection{\textbf{Training Categorical Model}}
To fit a Categorical distribution to a dataset, two factors are important: finding the right prediction space and the right training loss.
\\
\textbf{Discrete Tokenization in Action Modeling.} Frequently, robotic actions are continuous while categorical distributions predict distribution over discrete values. 
As a result, models with categorical distributions require ``tokenizing" the continuous actions into discrete ``action tokens''.
Often, this is done with an ad-hoc tokenizer binning every axis values into a certain number of bins, such as in~\cite{janner2021offline, brohan2022rt, brohan2023rt}.
Some models use a non-parametric algorithms such as k-means to tokenize the actions~\cite{shafiullah2022behavior, cui2022play, dadashi2021continuous}.
Later models~\cite{lee2024behavior} use more advanced generative methods, such as a VQ-VAE~\cite{van2017neural}, to tokenize the actions into discrete tokens.
\\
\textbf{Losses for Training.}
Quite frequently, Categorical models are trained with the \gls{ce} loss. Given a dataset $\gD:(\va_i, \vc_i)_{i=0}^N$, the \gls{ce} loss
\begin{align}
    \gL(\vtheta) = -\E_{j,\vc}\left[ \log \mC_{\vtheta}(\vc)_j \right],
\end{align}
is represented as the negative log-likelihood, where $j$ is class related the action in the dataset.
However, given the imbalance between different classes, some recent models~\cite{shafiullah2022behavior, cui2022play, lee2024behavior} use Focal loss~\cite{lin2017focal} instead
\begin{align}
    \gL(\vtheta) = -\E_{j,\vc}\left[ (1 - \mC_{\vtheta}(\vc)_j)^\gamma \log \mC_{\vtheta}(\vc)_j \right],
\end{align}
where $\gamma$ is a balancing hyperparameter. Focal loss is less sensitive for outliers and class imbalance, which can be important for tokenized action datasets.
\\
\textbf{On Action Value Maps.}
In the particular case of Action Value Maps, the action classes are projected to the pixel space of the input observation. 
The actions in the dataset $\va_j$ are first projected to a one-hot pixel map $H_j$ of the same shape of the visual observation $\vo_j$. The one-hot pixel map $H_j$ will set to one the pixel that relates to the action and zero otherwise. 
Then, the \gls{ce} loss is represented as
\begin{align}
    \gL(\vtheta) = -\E_{h, \vo \in \gD} [\log Q_{\vtheta}(h|\vo) ],
\end{align}
where $h$ denotes the positive pixel in the pixel map $H$. Despite using only the positive samples, the Softmax propagates the gradients to all the pixel-space, pushing the probability down to any pixel not being the positive one.

\begin{figure}[t]
    \centering
      \begin{minipage}[c]{0.39\textwidth}
    	\includegraphics[width=.99\textwidth]{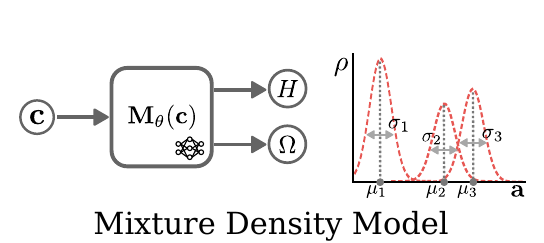}
    \end{minipage}
    \caption{A visual representation of \gls{mdm}. Given the context $\vc$ as input, \gls{mdm} $H, \Omega = \mM_{\vtheta}(\vc)$ outputs the parameters of a mixture density model, with $H : (\vmu_{0}, \vsigma_0 \dots, \vmu_{N}, \vsigma_N)$ the parameters of the models and $\Omega: (\omega_{0}, \dots, \omega_N)$ the weights for each model in the mixture.}
    \label{fig:mixture_density_model}
    \vspace{-.1cm}
\end{figure}

\subsection{Mixture Density Models}
\label{sec:models_mdm}

We call \gls{mdm}~\cite{bishop1994mixture, salimans2017pixelcnn++} to the set of generative models $H, \Omega = \mM_{\vtheta}(\vc)$ that take as input a context variable $\vc$ and output the parameters $H : (\veta_{0}, \dots, \veta_N)$ and the weights $\omega: (\omega_{0}, \dots, \omega_N)$ of a mixture density function. 
\begin{align}
    \rho_{\vtheta}(\va|\vc) = \sum_{k=0}^N \omega_k \rho(\va; \veta_k),
    \label{eq:mixture_density}
\end{align}
where each vector $\veta_k$ and scalar value $\omega_k>0$ represent the parameters and weight of the $k$ density model.
In practice, it is common to use 
\gls{gmm}~\cite{mandlekar2020learning, mandlekar2022matters, zhu2023viola, wan2023lotus, wang2023mimicplay} 
or Mixture of Logistic Models~\cite{salimans2017pixelcnn++, lynch2020learning, mees2022matters} to represent the mixture model.
Then the parameters $\veta_k = (\vmu_k, \vsigma_k)$ typically represent the mean $\vmu_k$, the standard deviation $\vsigma_k$, and $\omega_k$, the weight of each mode in the mixture model.

In the robotics literature, it is common to combine \gls{mdm} with additional generative models. In \cite{lynch2020learning, mandlekar2022matters, mees2022matters}, 
\gls{vae} is combined with \gls{mdm}.
Given a latent variable $\vz$ representing the high-level plan, a \gls{vae} decoder is trained to generate the parameters of a \gls{mdm} for the action space distribution. 
This differs from the classical use of \gls{vae}, where the output directly generates the action.

\subsubsection{\textbf{Main Application}}
A few works~\cite{mandlekar2022matters, zhu2023viola, wan2023lotus, mees2022matters, lynch2020learning} proposed representing visuomotor policies with \gls{mdm}, usually representing the action space as displacements in the end-effector.
Several of these works motivate the use of a mixture density model rather than an unimodal model in the representation of the policy to better capture the inherent multimodality in the demonstrations.

\subsubsection{\textbf{Training Mixture Density Model}}
Given a dataset $\gD:(\va_i, \vc_i)_{i=0}^N$, \gls{mdm} are trained by minimizing the negative log-likelihood of the learned model
\begin{align}
    \gL(\vtheta) = -\E_{\va,\vc\sim\gD}\left[\log \rho_{\vtheta}(\va|\vc) \right]
\end{align}
with $\rho_{\vtheta}(\va|\vc)$ (See \Cref{eq:mixture_density}) being the density model parameterized with the output of the learned model $H, \Omega = \mM_{\vtheta}(\vc)$.

\section{Integrating Generative Models into Robotics}
\label{sec:applications}

\begin{figure*}[t]
    \centering
      \begin{minipage}[c]{0.99\textwidth}
    	\includegraphics[width=.99\textwidth]{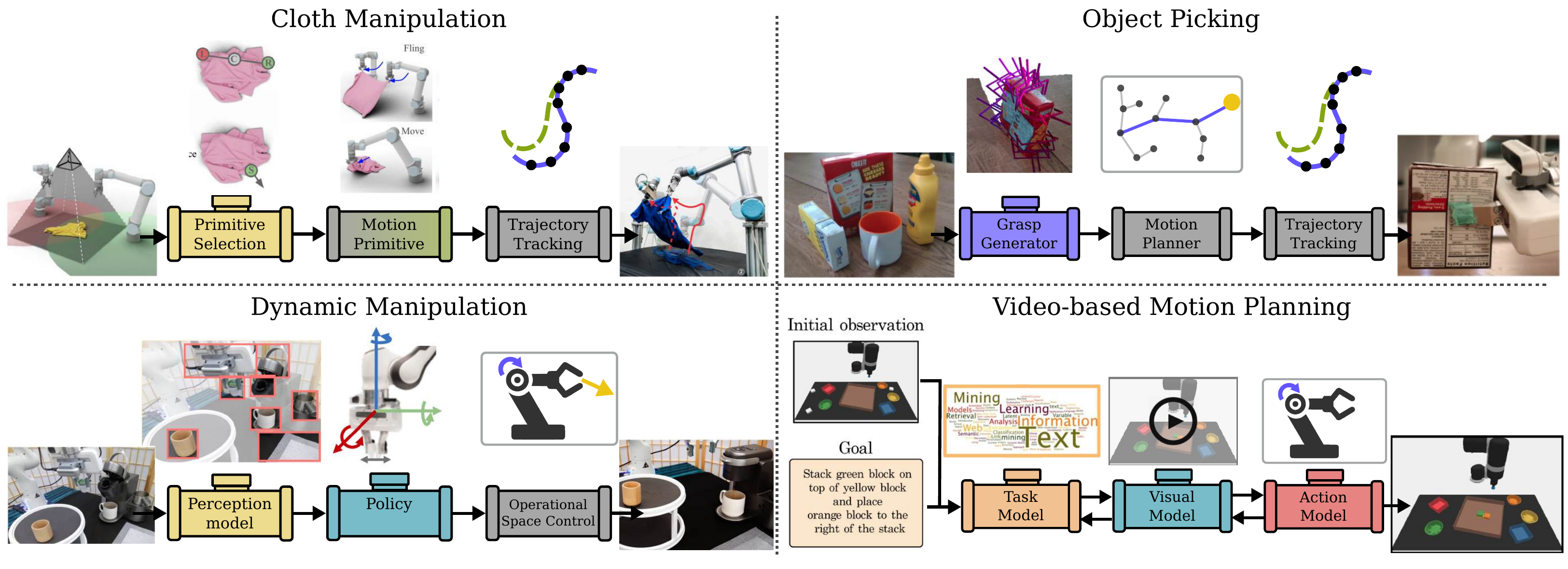}
    \end{minipage}
    \caption{\small{Visual representation of different approaches to apply \gls{dgm} in robotics tasks. Colored: Learned models, Grey: Predefined models. \textbf{(a)} Cloth Manipulation. Given a set of motion primitives, an Action Value Map selects the primitive and the parameters of the primitive~\cite{ha2022flingbot}. \textbf{(b)} Object Picking. An SE(3) pose generative model generates a target pose to grasp an object and a motion planner generates the path to reach the grasp~\cite{mousavian20196}. \textbf{(c)} Visuo-Motor Policy. Given an image as input, a visuomotor policy generates end-effector actions. Then, an Operational Space Controller maps the action to the configuration space~\cite{zhu2023viola}. \textbf{(d)} Video Planning. A \gls{llm} generates a plan in text. The text generates a video of the substeps. Then, a goal-conditioned policy generates robot actions conditioned on generated images~\cite{ajay2023compositional}.}}
    \label{fig:sec4_generative_apps}
    \vspace{-.5cm}
\end{figure*}

In this section, we look at different design strategies for integrating \gls{dgm} into robotics.
The number of robotic tasks in which \gls{dgm} has been applied is vast; pick and place tasks~\cite{murali20206, weng2023neural, urain2022se3dif}, cloth manipulation~\cite{weng2022fabricflownet, ha2022flingbot}, scene rearrangement~\cite{liu2022structformer, liu2022structdiffusion}, food preparation~\cite{chi2023diffusion, zhu2023viola, zhao2023learning}.

A common strategy for using \gls{dgm} in this wide range of tasks is to integrate the learned \gls{dgm} into a larger framework.
The learned model is typically combined with other predefined or learned components, such as perception modules, motion primitives, task and motion planners, or controllers, creating a synergy that exploits the strengths of both learned and predefined components.
The \gls{dgm}s are tasked with addressing the components of the problem that are difficult to model conventionally, while the predefined elements handle aspects that are easier to define. 
This combination enhances the system's ability to tackle complex tasks by leveraging the predictive power and adaptability of the \gls{dgm}s.

Due to the variety of tasks and the flexibility in which \gls{dgm} can be combined with other components, there are a wide variety of ways in which \gls{dgm} have been integrated into robotics problems. We present some possible combinations in \Cref{fig:sec4_generative_apps}.
In the following, we present some of the most common strategies that the robotics community has used to integrate \gls{dgm} to solve robotics tasks. We classify the strategies based on the element that \gls{dgm} generates.

\subsection{Generating End Effector Target Poses}
End-effector 6D (position and orientation) poses are among the most common elements generated with \gls{dgm}~\cite{mousavian20196, shridhar2023perceiver, gervet2023act3d, urain2022se3dif, murali20206, breyer2021volumetric, goyal2023rvt, simeonov2022neural, simeonov2023shelving}. 
The generated poses were used as target grasping poses of an object in the scene~\cite{mousavian20196, murali20206, simeonov2022neural, urain2022se3dif}, as target placing poses for the objects~\cite{simeonov2023shelving}, or as an action of a policy~\cite{shridhar2023perceiver, gervet2023act3d, goyal2023rvt}.

\textbf{When to use.} SE(3) pose-based \gls{dgm} have been particularly successful in capturing particularly \textit{relevant target locations in robot tasks}. For example, in a pick and place task, the SE(3) pose model will inform the desirable grasping pose and the desirable placing pose. In a drawer opening task, the SE(3) pose might inform the desirable pose for grasping the drawer handle and where to move to properly open the drawer.

\textbf{How to use.} 
The SE(3) pose generative models typically exploit the symmetries between the scene and the poses, leading to improved generalization in novel scenarios. More details can be found in \Cref{sec:generalization}.
The integration of the output of the \gls{dgm} in robotics applications depends on the task.
\\
In \texttt{offline tasks}, where the robot behavior is computed offline and executed in open-loop, it is common to combine the \gls{dgm} with motion planning modules. In \cite{mousavian20196, murali20206, jiang2021synergies, jauhri2023learning}, the generated SE(3) pose is used as the target pose in a motion planning problem. Then, sampling-based~\cite{kavraki1996probabilistic, lavalle1998rapidly} or optimization-based~\cite{ratliff2009chomp, kalakrishnan2011stomp} motion planners are used to generate the configuration space trajectories. In \cite{urain2022se3dif, weng2023neural}, the learned \gls{dgm} is integrated directly into the motion planning problem. Instead of sampling a pose, since the learned model is a \gls{ebm}, the model is integrated as an additional cost function.
\\
In \texttt{online tasks}, where the generative model is used as a policy, the proposed solutions depend on the horizon of the output.
In \cite{shridhar2023perceiver, gervet2023act3d, goyal2023rvt, ze2023gnfactor, ke20243d}, the generated SE(3) pose is used as a target pose in a motion planning problem. Then, since the problem to be solved is sequential, the system sequentially generates new trajectories as the robot reaches the previous target pose.
Some recent works~\cite{xian2023chaineddiffuser, ma2024hierarchical}, have proposed to replace the motion planner with a trajectory diffusion model conditioned on the generated SE(3) pose.
To have a higher control frequency, some works~\cite{urain2022manifold_svf} have proposed using operational space controllers~\cite{nakanishi2008operational, khatib1987unified} instead of a motion planner, usually considering a shorter horizon SE(3) target pose.

\subsection{Trajectories}
In recent years, \gls{dgm} have shown the ability to generate high-dimensional data. 
This has led the research community to learn \gls{dgm} that generate complete trajectories~\cite{janner2022planning, chi2023diffusion, carvalho2023motion, reuss2023goal, zhao2023learning, jiang2023motiondiffuser, huang2023diffusion}, rather than single-step actions.

\textbf{When to use.} Directly generating the entire trajectory can have several advantages over generating single target poses. 
First, trajectory generation allows the generation of \textit{complex dynamic motions} (pouring tomato into a pizza~\cite{chi2023diffusion}, inserting a battery~\cite{zhao2023learning}) in contrast to motion planning based methods that are usually limited to reaching tasks.
Second, trajectory generation may lead to \textit{smaller covariate shift errors} in contrast to single step generation. Because we generate trajectories, the execution of a long-horizon task might require fewer sequential decision steps, leading to smaller accumulated errors~\cite{zhao2023learning}.

\textbf{How to use.} The most common applications of trajectory \gls{dgm} are integrated into offline motion planners~\cite{ratliff2009chomp, mukadam2016gaussian} or used as receding horizon controllers~\cite{mayne1988receding, williams2017model, chua2018deep}.
\\
In \texttt{offline motion planning}, a number of works~\cite{janner2022planning, carvalho2023motion, huang2023diffusion} compose the trajectory \gls{dgm} with heuristic cost or reward functions.
Since the generative model is a \gls{dm}, the sampling process is conditioned on additional reward/cost functions to generate trajectories that satisfy additional objectives such as reaching a goal~\cite{janner2022planning} or avoiding obstacles~\cite{carvalho2023motion}.
This approach is similar to classifier-guided generation~\cite{dhariwal2021diffusion}, where a diffusion model for image generation is conditioned with additional classifiers.
Other works have composed multiple trajectory \gls{dgm} to generate long horizon trajectories~\cite{mishra2023generative} by sequential composition.
\\
In \texttt{receding horizon control} problems, a trajectory \gls{dgm} iteratively generates new trajectories adapted to changes in the scene. 
In \cite{chi2023diffusion}, a \gls{dm} is used to generate future action trajectories.
The authors assert the importance of predicting a sequence of actions to overcome possible latency gaps caused by image processing, policy inference, and network delays.
Instead of using \gls{dm}, \cite{zhao2023learning} uses a \gls{vae} to generate action trajectories. To generate the whole trajectory, they use an autoregressive generation approach, in which the action for a single step is generated at each step. 
To ensure smooth behavior, the authors propose a temporal ensemble of the generated actions with weighted averaging.

Once the desired trajectories are computed, a trajectory tracking controller is applied to move the robot along the generated path.

\subsection{Generating End-Effector Displacement}
One of the most common action spaces for policy learning is end-effector displacement~\cite{mandlekar2022matters, mees2022matters, brohan2022rt, zhu2023viola, team2023octo}. End Effector Displacements refer to 6 DoF changes in both the position and orientation of the robot's end effector that are directly related to the velocity in the end effector.

\textbf{When to use.} Displacement \gls{dgm} have been commonly used as \textit{control policies}. The low dimension of the generated variable allows a high frequency generation in contrast to trajectory \gls{dgm}. This allows the robot to adapt quickly to changes in the environment, leading to \textit{highly reactive} policies. Additionally, in combination with an in-hand camera, we can build policies that act locally, providing a high degree of generalization.
For example, if the in-hand camera observes an apple to be picked, given the actions are generated wrt. the end-effector, the robot uses only relative information and adapts its behavior to novel situations as long as the relative position of the end-effector and the apple remains the same.

\textbf{How to use.}
End-effector displacement \gls{dgm} are integrated into feedback controllers. The output of the \gls{dgm} can be integrated into impedance controllers informing on the desirable motion direction.
In visuomotor policies, it is common to represent the displacement in the robot's wrist camera frame. 
Given as observation the camera images, this allows representing local policies that could generalize its performance to scenes in which the scene looks similar locally from the camera view~\cite{mees2022matters}.
Rather than conditioning the policy on the last observation, several works~\cite{mandlekar2022matters, shafiullah2022behavior, mees2022matters} have found it useful to condition the model on a history of observations by using LSTM or RNN networks. This allows the policy to encode relevant features that might not be possible to extract from the last observation.
Different \gls{dgm} have been applied to capture end-effector displacements, from \gls{mdm}~\cite{mandlekar2022matters, zhu2023viola, mees2022matters}, categorical distributions~\cite{brohan2022rt, brohan2023rt}, \gls{ebm}~\cite{florence2022implicit} to \gls{dm}~\cite{team2023octo}.

\subsection{Scene Arrangements}
A set of works proposes generating desirable target scenes~\cite{liu2022structformer, liu2022structdiffusion, mishra2023reorientdiff, kapelyukh2022dall, simeonov2023shelving, yang2023compositional, gkanatsios2023energy}. The target scene is represented as a set of SE(3) poses for different objects.

\textbf{When to use.}
Scene arrangement \gls{dgm} are commonly applied to generate desirable placing poses for a set of objects in a scene. Given, we know the different objects in the scene, a set of works considers the generation of the placing poses of those objects given text commands informing on the desirable scene arrangement~\cite{liu2022structformer, gkanatsios2023energy, kapelyukh2022dall, liu2022structdiffusion, mishra2023reorientdiff}. For example, given a text command "\textit{Set the table for dinner}", the scene arrangement \gls{dgm} will generate a set of placing poses for dishes, glasses, and cutlery.

\textbf{How to use.}
Scene arrangement \gls{dgm} are commonly integrated into robot tasks with task and motion planners. Given a generated set of placing poses for a set of objects in the scene, the task and motion planner decides the order in which each object should be placed and the robot motion to pick and place each object.
This type of model assumes access to some form of object representation.
In \cite{gkanatsios2023energy}, the bounding box of the different object's of interest is extracted from an image.
In \cite{kapelyukh2022dall}, semantic masking is applied with Mask R-CNN~\cite{he2017mask}.
In \cite{liu2022structdiffusion}, the pointcloud of the object's in the scene is cropped to represent the different objects.
Several works have explored composing multiple scene arranging models to generate complex arrangements~\cite{gkanatsios2023energy, yang2023compositional}. 
In \cite{gkanatsios2023energy} multiple \gls{ebm} are composed with different objectives (such as place the fruit in circle \gls{ebm} and place the fruits on the plate \gls{ebm}). 
In \cite{yang2023compositional}, the arrangement of a set of objects in the scene is framed as a constraint satisfaction problem. The work composes a set of \gls{dm} representing the relative pose of the objects between each other.



\section{Generalizing outside data distributions}
\label{sec:generalization}
The problem of generalization in generative modeling refers to the ability of a \gls{dgm} to produce high-quality, meaningful samples beyond the training dataset.
In the particular case of robotics, given a generative model $\rho_{\vtheta}(\va|\vc)$ trained on a dataset $\gD:\{\va_n,\vc_n\}_{n=1}^N$, the goal is to generate useful actions $\va$ for contexts $\vc \notin \gD$ that are not part of the dataset.

Achieving high generalization capabilities requires the selection of smart architectural choices that enhance the agent's generalization capabilities.
We group the strategies in three main categories:
\\
\textbf{Composition.} 
Rather than learning a monolithic policy, several researchers have explored learning individual behavior modules that can later be composed to generate complex behaviors. This composition can be both parallel (combining the behaviors together) and sequential (for generating long-horizon tasks). The composition of simple modules allows the formation of complex models that can generalize to new tasks not seen in the demonstrations~\cite{du2024compositional}.
\\
\textbf{Feature Selection from Observations.} 
With a small dataset, the robot is expected to learn spurious correlations between observations and actions, in part due to the high dimensionality of the observations (images, tactile signals). To alleviate this problem, several researchers have explored the problem of reducing the observations to informative features, thus improving the system's ability to learn meaningful correlations.
\\
\textbf{Observation-Action Symmetries.}
In visuomotor policies, visual observations and actions are typically represented in different spaces. Given this mismatch, it is difficult for the policy to learn the correct relations between observations and actions, leading to poor generalization when observations are changed. To reduce this mismatch, a large line of research explores the problem of representing both visual observations and actions in a common space. In this context, some approaches propose mapping actions to pixel space or representing both visual observations and actions in 3D space.

In the following, we elaborate on these three approaches and show how they have been useful for robotics problems.

\begin{figure}[t]
    \centering
      \begin{minipage}[c]{0.49\textwidth}
    	\includegraphics[width=.99\textwidth]{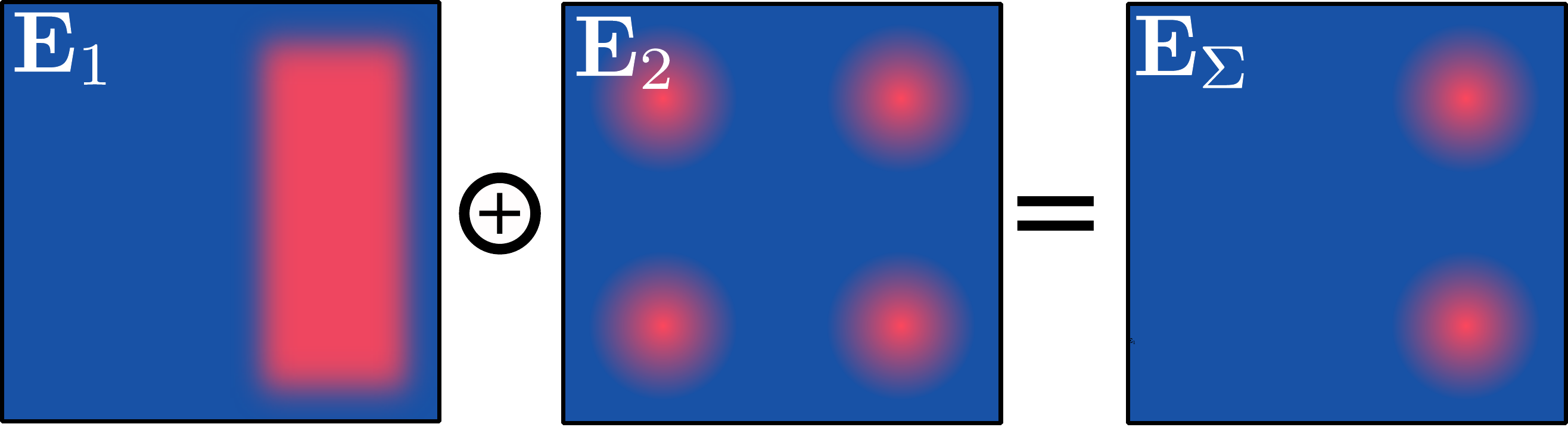}
    \end{minipage}
    \caption{Visual representation of energy composition. Due to the implicit nature of \gls{ebm}, the distribution of the models can be composed to satisfy multiple objectives.}
    \label{fig:energy_composition}
    \vspace{-.5cm}
\end{figure}

\begin{figure*}[t]
    \centering
      \begin{minipage}[c]{0.99\textwidth}
    	\includegraphics[width=.99\textwidth]{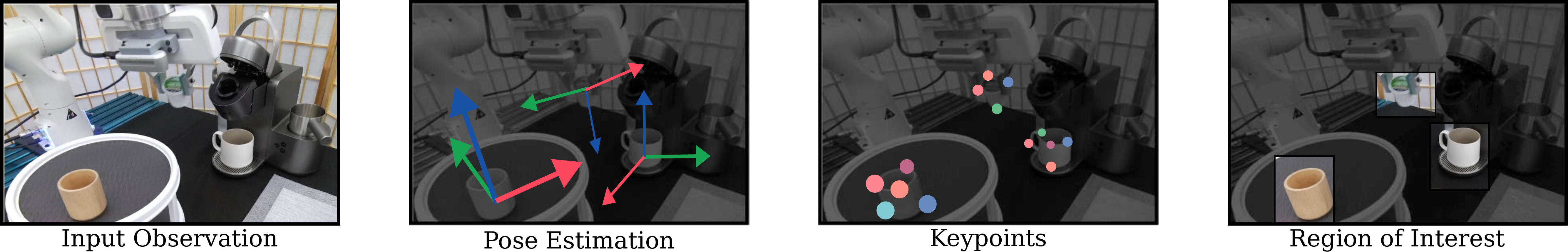}
    \end{minipage}
    \caption{Given the high-dimensionality of the visual contexts, different works explore how to extract informative features from the context by an information bottleneck. Among the options, we can extract the pose of the relevant objects, extract relevant keypoints or extract the regions of interest in the image.}
    \label{fig:bottlenecks}
    \vspace{-.5cm}
\end{figure*}

\subsection{Composition}
Many robot tasks can be decomposed into a composition of simpler subtasks.
Consider, for instance, the problem of selecting a grasping pose to pick an object.
The selection of the pose can be decomposed into the satisfaction of multiple simpler objectives: satisfying the robot's joint limits, avoiding collisions in the scene, and constructing a geometrically consistent grasp pose on an object.
While learning a monolithic \gls{dgm} to take into consideration all the objectives would require gathering a set of demonstrations satisfying these constraints, we can solve this task in a zero-shot manner by directly combining simpler models that capture each simpler objective. This modular approach allows us to learn a set of simple objectives that can combined at test-time to solve complex new unseen tasks~\cite{du2024compositional}.


Formally, given a generative model $\rho(\va|\vc_1)$ which generates actions conditioned on a objective $\vc_1$ and another generative model $\rho(\va|\vc_2)$ which generates actions conditioned on a objective $\vc_2$, the composition
\begin{align}
    \rho(\va|\vc_1, \vc_2) \propto \rho(\va|\vc_1)\rho(\va|\vc_2),
\end{align}
leads to a generative model $\rho(\va|\vc_1, \vc_2)$ that will generate actions that are likely for both $\vc_1$ and $\vc_2$ objective~\cite{du2020compositional}.
Alternatively, given a discriminative model $\rho(\vc_2|\va)$, the composition
\begin{align}
    \rho(\va|\vc_1, \vc_2) \propto \rho(\va|\vc_1)\rho(\vc_2|\va),
\end{align}
also allows us to construct a generative model  $\rho(\va|\vc_1, \vc_2)$ through Bayes rule. Thus, composing multiple probabilistic models allows for the construction of samples that satisfy multiple objectives.

In \cite{murali20206}, two models are composed to generate grasp poses that are both valid to grasp any object and also collision-free.
First, a \gls{vae}-based grasp pose generative model~\cite{mousavian20196} generates a set of grasp candidates. Then,
a discriminative model evaluates collisions in the scene for the grasp candidates. 

Due to their implicit nature, \gls{ebm}s and \gls{dm}s have been extensively integrated for composable sampling.
One early example in \gls{ebm}s is in \cite{Du2019ModelBP} which illustrates how combining a trajectory-level EBM with a reward function can implement model-based planning and generate high-reward trajectories. In \cite{janner2022planning}, this approach is applied to diffusion models, where a generalistic trajectory-level diffusion model is combined with a reward function to generate high-reward samples among the demonstrations.

A set of works~\cite{urain2022se3dif, saha2023edmp, carvalho2023motion, yang2023compositional, luo2024potential} have illustrated in various forms how multiple cost or constraint functions are combined to define an optimization problem over trajectories. For example, in \cite{urain2022se3dif}, a diffusion in SE(3) representing valid grasp poses is combined with collision-avoidance, trajectory smoothness, or robot joint limits cost to generate collision-free pick and place trajectories.

Compositionality can also be applied sequentially along a temporal axis  to generate trajectories for solving long-horizon tasks. In \cite{mishra2023generative, mishra2024generative, du2023video}, diffusion models are composed sequentially to solve long-horizon manipulation tasks. The authors propose learning diffusion models for individual skills and then sample trajectories by chaining skills and sampling from the joint distribution.

Composability have been also explored to induce generalization in the object class~\cite{liu2023composable} or among sensor modalities~\cite{wang2024poco}. In \cite{liu2023composable}, the motion of the tools to solve a given task is generated by diffusion models. The work proposed segmenting the tool into different sections~(handle, body, rim) and composing the diffusion models defined for the different parts. This decoupling induces a generalization to novel objects with different shapes and sizes. In \cite{wang2024poco}, several learned models are combined in which each model might depend on a different sensor modality from images, tactile or pointcloud. This composition allows the policy to learn sensor-specialized skills and combine them afterwards.

Finally, composability can be applied to foundation models trained on separate sources of Internet knowledge~\cite{ajay2023compositional,du2023video} to combine information across each model. In ~\cite{ajay2023compositional}, a large language model capturing high-level information is combined with a video model capturing low-level information and an egocentric action model capturing action information. By composing all three models together, the model can in a zero-shot manner solve long-horizon tasks by integrating the knowledge across all three models. This composition is extended in~\cite{du2023video}, where planning between a vision-language and video model is used to construct long horizon plans.

\subsection{Extracting the informative features from the perception}
Given the high amount of information in the visual observations $\vc$, to properly solve a robotics task, we might require to apply some form of representation learning to focus on the meaningful features to solve the tasks.
For example, due to the limited training data, end-to-end visuomotor policies are likely to falsely associate actions with task-irrelevant visual factors, leading to poor generalization in new situations~\cite{zhu2023viola, wang2021generalization}.
In contrast, with a proper representation learning approach, the robot might learn meaningful features for generalization beyond the demonstrations.

Consider a language-conditioned policy trained on demonstrations that include specific text commands, such as "open a drawer".
Crucially, the ability of these models to generalize to semantically similar but lexically distinct commands, such as "pull-out a drawer" without direct training on such commands, represents a significant advancement in generalization.
Another example might be in an image-conditioned policy. Given a learned model, the robot should be able to generalize its behavior to scenes in which distractors might appear or objects are located in novel places.

This generalization is facilitated by learning an encoder $\vz = \gE(\vc)$ that is capable of producing latent representations $\vz$ that capture the relevant features to solve the robotics tasks.

Related to \textbf{visual contexts}, a common approach is to extract some form of \textbf{object-centric features} from the images, usually related to the location of the objects.
\\
A classical approach is to pre-train a \texttt{pose estimation} model, that will transform a visual input into the position $\vp \in \RR^3$ and orientation $R \in SO(3)$ of the object of interest~\cite{yoon2003real, sahin2018category, deng2020self}.
Nevertheless, 
as pointed out in \cite{manuelli2019kpam}, a category-level pose estimation can be ambiguous under large intra-category shape variations. For example, knowing the pose of a coffee mug might not be enough to successfully hang it on a rack, as different coffee mugs might have different handles shapes or handle locations.
\\
Alternatively, a set of works has proposed extracting a set of \texttt{key points} from the image~\cite{levine2016end, manuelli2019kpam, manuelli2020keypoints, sieb2020graph, kulkarni2019unsupervised, qin2020keto, wang2021generalization}. For example, in \cite{manuelli2019kpam}, a 3D keypoint detection network transforms an RGB-D image into a set of 3D keypoints $\mP=\{\vp_i\}_{i=1}^N \in \RR^{N\times 3}$, where $N$ is the number of keypoints.
In contrast with only extracting the pose, several key points could inform about the shape of the object of interest.
\\
A more general approach is to extract a \texttt{set of cropped images} through bounding boxes~\cite{devin2018deep, wang2019deep, zhu2023viola}.
Given an RGB image as input, the encoder outputs a set of Regions of Interest (RoI) represented by bounding box locations $\vp_i \in \RR^4$ (pixel locations to construct the bounding box) and the cropped images $\mI_i^{\text{crop}}$ by the given pixel locations.
While \cite{devin2018deep, wang2019deep} consider a category-level training to extract the bounding boxes, in \cite{zhu2023viola}, a general pre-trained Region Proposal Network (RPN)~\cite{ren2015faster} is used to extract the cropped images. Then, a Transformed policy, sets the attention on task-relevant cropped images.
\\
An alternative approach to get the cropper images is through \texttt{segmentation masks}~\cite{heravi2023visuomotor, zhu2023learning, wu2021apex}.
For example, in \cite{heravi2023visuomotor}, Slot Attention~\cite{locatello2020object} is applied to extract the different segmentation masks of the objects in the scene unsupervised.
In \cite{zhu2023learning}, it is proposed to provide both demonstrations and scribbles on the important objects to pay attention to. Then, an interactive segmentation model~\cite{cheng2021modular} generates the segmentation mask of the desirable objects.
\\
A recent line of research explores using \texttt{language conditioned semantic features} from images \cite{shafiullah2022clip, shen2023distilled, shridhar2022cliport, huang2023visual}. Given a language command, the model highlights the semantically most aligned features allowing the robot's behavior to focus mostly on them. This relation between language and vision inputs is commonly obtained by computing the cosine distance between the CLIP features~\cite{radford2021learning}. This approach is particularly relevant for robotics as it allows to exploit the pre-trained vision language models in an efficient way.

In a different direction, a few works have explored how to integrate \textbf{tactile information} for robot manipulation.
A common strategy has been to \textbf{reconstruct 3D shapes from tactile}~\cite{wang20183d, suresh2023neural, smith20203d, suresh2022shapemap, chen2023sliding, yuan2023robot}.
In \cite{wang20183d}, a vision-based predicted 3D shape, represented by a voxel-grid is updated with multiple touches on the object. The tactile information in combination with the location of the sensor is converted into occupied voxel information to ground it the 3D space.
In \cite{suresh2023neural}, the shape of the object is reconstructed into a Neural SDF, while manipulating the object. Given the object's orientation and pose is changed, the work combines a pose estimation with a shape reconstruction objective.
In \cite{yuan2023robot}, the tactile signals are represented as a 3D pointcloud. Given binary sensors, the authors transform the signal into a 3D pointcloud if the sensor is in contact with an object.


\subsection{Exploiting Symmetries between Perception and Action}
Multiple robot tasks have inherent symmetries.
Consider, for instance, a top-view picking problem. Given a demonstration of the desired grasp pose to pick an apple; if the apple is moved 10 centimeters, the desired grasp pose should similarly move 10 centimeters. Thus, building policies that exploit this symmetry will induce important generalization.

Representing both perception and action in a shared space has shown important results in this direction~\cite{morrison2018closing, zeng2017robotic}.
Given both (action and observation) are represented in the same space, the generative model exploits the spatial structure and allows building architectures that contain spatial symmetries, such as translation equivariance~\cite{wu2020spatial, zeng2021transporter}.

\begin{figure}[t]
    \centering
      \begin{minipage}[c]{0.35\textwidth}
    	\includegraphics[width=.99\textwidth]{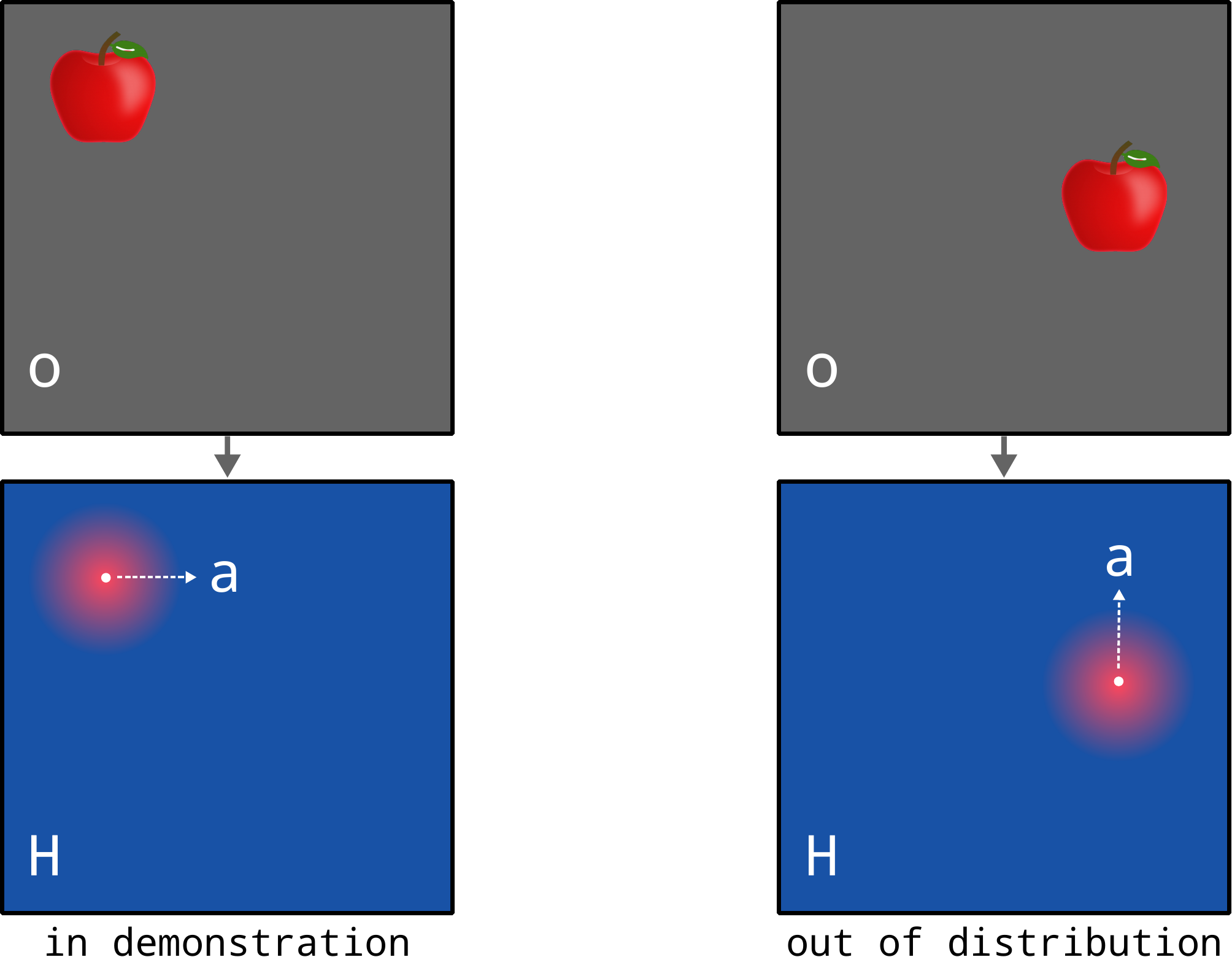}
    \end{minipage}
    \caption{Action Value Maps contain spatial symmetries by projecting the action to the pixel space. Given the value map $\mH$ is computed by local features in the input image $\vo$, a translation of the pixels, leads to a translation in the value map.
    Figure inspired by \cite{zeng2021transporter}.}
    \label{fig:action_map_symmetry}
    \vspace{-.5cm}
\end{figure}

A common policy architecture to ground the actions into the perception is known as Action Value Map~\cite{wu2020spatial, ha2022flingbot} or Affordance Map~\cite{zeng2017robotic, shridhar2022cliport} (See \Cref{fig:category_model}).
Consider, the top-view picking problem. Given a visual observation $\vo$ of the apple to pick, an Action Value Map $\rho(\va|\vo)$ will learn to place a high probability on the pixels around the apple (given the action is grounded in the pixel space) and a low probability on the rest of the space. Then, in inference time, even if the apple is translated, the action distribution will similarly translate to the region where the apple is.
We visualize it \Cref{fig:action_map_symmetry}.

This model type has been particularly successful on \textbf{top-view manipulation tasks}.
One of the first applications of Affordance Models was for grasp pose generation in bin-picking problems~\cite{zeng2017robotic}.
Given an image as input, the model outputs a value map in the pixel space, representing the quality of all 2D locations to pick an object via suction.
To consider the orientation of a parallel gripper, \cite{zeng2017robotic} rotates the observation image by 16 different angles and generate 16 value maps, one per rotated image. Each rotated value map is used as a possible orientation candidate for the grasp.
\cite{morrison2018closing} instead, generates an additional value map informing about the optimal orientation per pixel.
In \cite{zakka2020form2fit, zeng2021transporter, shridhar2022cliport}, Affordance models are extended to pick and place problems. In \cite{zakka2020form2fit}, the correlation between the picking and the placing actions is induced through a matching module that infers the correspondence between possible picking objects and possible placing locations.
In \cite{zeng2021transporter}, the placing value map is conditioned on a crop image along a selected picking pixel.
In \cite{shridhar2022cliport}, Transporter Networks~\cite{zeng2021transporter} are extended to consider language goal $\vg$ in addition to the visual observations $Q_{\vtheta}(\vo,\vg)$.
In \cite{huang2022equivariant}, Transporter Networks are extended with Equivariant networks. This leads to not only translation equivariance but also rotation equivariance.

This type of models have been particularly useful for deformable objects~\cite{seita2021learning, ha2022flingbot, avigal2022speedfolding, weng2022fabricflownet}.
In \cite{seita2021learning} the problem of rearranging a deformable object is solved as a sequence of pick and place actions. In their work, Transporter Networks are extended to learn a goal-conditioned pick and place policy.
In \cite{ha2022flingbot}, a bimanual robot is trained for cloth manipulation. 
The Affordance Model is trained to select the parameters of a Flinging policy. Similarly to \cite{zeng2017robotic} the image is rotated to consider different possible grasping orientations. Addtionally, the image is scaled to different sizes to parameterize the distance between both manipulators when flinging.

Beyond Pick and Place or deformable object manipulation, in \cite{zeng2020tossingbot}, an Affordance model is trained to throw objects. An Affordance model first selects the place to pick an object, then a throwing velocity module assign a desirable throwing velocity to that pixel. In \cite{zeng2018learning}, a policy is learned to tidy up a table. Given two primitives (push and pick), an Affordance Model is trained per each primitive and the most likely action is selected among all value maps.
Finally, in \cite{wu2020spatial}, Affordance models were applied in a mobile navigation task, in which the robot needs to manipulate a set of objects.

Grounding perception and action have been also explored for \textbf{6-DoF manipulation}.
In \cite{james2022coarse, shridhar2023perceiver, grotz2024peract2}, Action Value Maps are extended to a voxel grid space. Given as input a voxel representing a 3D space, the action space is defined as a categorical distribution along the voxels, where each voxel represents a target 3D location to move the end-effector.
To generate the voxel-grid with meaningful semantic information, \cite{ze2023gnfactor} propose constructing the voxel-grid combining Neural Radiance Fields~\cite{mildenhall2021nerf} and Stable Diffusion~\cite{rombach2022high}.
Voxel-based network are usually computationally demanding. To tackle it, RVT~\cite{goyal2023rvt, goyal2024rvt} instead proposes projecting the problem into multiple image-level Action Value Maps. 
Given multiple viewpoints, RVT proposes generating an Action Value Map for each view and then, sample the action through an optimization over all views.
Similarly \cite{lin2023mira} also projects the 6-DoF manipulation problem to an image Action Value Map. In this case, the authors solve an optimization problem among multiple viewpoints to select the one that provides the best top view for solving the task.

Instead of using a voxel grid, a set of works have explored using point cloud representations in which the action is directly projected in the point cloud itself~\cite{mo2021where2act, wu2021vat, geng2023rlafford, wang2022adaafford, zhao2022dualafford, mo2022o2o}. Similar to image-based value maps, these approaches represent a categorical distribution along the points in the pointcloud, where each point is a possible action and the model outputs a probability over all the points. A particular case can be found in \cite{gervet2023act3d}, where the action can be represented in any point in the 3D space, given a pointcloud as observation.

A limitation of Action Value Maps is that do not scale to large action spaces such as trajectories. To represent higher dimensional action spaces, several works have explored integrating observation-action symmetries in \gls{dm}. \cite{vosylius2024render, shridhar2024generative} compute the denoising step by projecting the action candidate to a set of camera views. Alternatively, \cite{xian2023chaineddiffuser, ke20243d} first build a featurized 3D pointcloud scene and denoise the actions directly in the 3D space by computing the relative distance between observations and actions.









\section{Future Research Directions}
\label{sec:research_challenges}
Despite the successful deployment of \gls{lfd} in several robot tasks, there are several open research challenges.
We consider three main pillars will drive the future research in \gls{lfd}:
\begin{itemize}
    \item How do we solve long-horizon tasks?
    \item How do we obtain large amounts of data to train \gls{dgm}?, and how do we learn from them?
    \item How do we guarantee policies to generalize to novel goals and novel scenes?.
\end{itemize}
In the following, we present a set of future research directions to apply \gls{lfd} methods to solve robotics tasks.
\\
\\
\textbf{Robot policies for long horizon tasks.}
Long-horizon tasks are usually solved through task and motion planning algorithms.
These approaches are usually crafted for specific applications and do not generalize to any possible task.
On the other hand, learning-based policies are usually limited to short-horizon skills.
Learning policies that are able to solve any type of long-horizon task is an open research question.
A promising direction is the combination of \gls{llm} for high-level task planning with low-level, short-horizon robot skills~\cite{ajay2023compositional, ahn2022can}.
Nevertheless, properly exploiting the outputs of the \gls{llm} for task generation will require a proper \textit{grounding of the language commands with robot actions}.
\\
\textbf{Learning from video demonstrations.}
Teleoperation data is one of the most common approaches to demonstrate to the robots how to behave.
Nevertheless, collecting large amounts of teleoperated data is costly.
On the contrary, the internet is full of videos of humans performing all sorts of tasks. These videos are an important source of data to teach robots the desirable behavior to solve any sort of task.
Several strategies have been explored, from extracting informative features from video~\cite{peng2020learning, qian2024pianomime}, learning directly rewards from the videos~\cite{fan2022minedojo, ma2022vip}, or learning video generative models~\cite{du2023video, du2023learning}.
Among the different challenges to properly learn from videos are solving the \textit{embodiment mismatch between the human and the robot}, \textit{lack of direct action data}, or the \textit{mismatch between training and testing environments}. 
\\
\textbf{Learning from synthetic data.}
Given the difficulty of collecting real robot data, physics simulators emerge as a possible approach to generate large amounts of data.
In this direction, there have been several works~\cite{james2020rlbench, liu2024libero, gu2023maniskill2, mandlekar2023mimicgen, nasiriany2024robocasa} that built benchmarks in simulation and provide pipelines for synthetic data generation.
Nevertheless, deploying real robot policies trained on synthetic data requires properly addressing the \textit{sim-to-real gap}. 
\\
\textbf{Learning from online interaction.}
Given the high variability of the possible scenes a robot could encounter, learning a generalist single policy from an offline dataset for all possible tasks is unfeasible. Instead, an important research direction proposes training policies for new tasks by allowing the robot to interact with the environment in which it will be deployed~\cite{celemin2022interactive}. This requires the robot to explore different possible behaviors to find those that are most suitable for the task in deployment. However, \textit{the way the robot explores and learns to solve new tasks is critical for efficient learning of new policies and is an important direction of future research.}
\\
\textbf{Generalization.}
Even if the models are trained on large amounts of data, the robot will likely encounter situations that were not in the dataset.
Thus, the generative model should be capable of generalization, generating good actions in unseen situations.
As shown in \Cref{sec:generalization}, a proper selection of inductive biases can promote generalization capabilities.
Despite some interesting properties, current generative models have not shown yet powerful generalization capabilities and additional exploration on \textit{structured priors for generalization} is an important direction of future work. In addition, \textit{integrating internet knowledge} can be additional source of generalization performance. Existing foundation models capture rich sources of information from the internet that a robot policy can exploit to generalize to new settings.
Finally, structure in terms of 3D geometry can further help in the grounding and aggregating of semantic information for robot policies, leading to better generalization. In this regard, \textit{3D Feature fields}~\cite{gervet2023act3d, shafiullah2022clip, shen2023distilled} are a direction to represent semantic information and the robot actions in common space.

\bibliographystyle{IEEEtran}
\bibliography{bibliography.bib}

\printglossaries

\end{document}